\documentclass[preprint,5p,times,authoryear,twocolumn]{elsarticle}
\usepackage{amssymb}
\usepackage{amsmath}
\usepackage{times}
\usepackage{soul}
\usepackage{url}
\usepackage[hidelinks]{hyperref}
\usepackage[utf8]{inputenc}
\usepackage[small]{caption}
\usepackage{graphicx}
\usepackage{amsmath}
\usepackage{amsthm}
\usepackage{booktabs}
\usepackage{algorithm}
\usepackage{algorithmic}
\usepackage{multirow}%
\usepackage{amsfonts}%
\usepackage{tikz}%
\usetikzlibrary{trees}%
\usepackage{tabularx}%
\usepackage{amssymb}%
\usepackage{arydshln}%
\usepackage{makecell}%
\usepackage[most]{tcolorbox}%
\usepackage{xcolor}%
\usepackage{colortbl}%
\usepackage{bm}%
\usepackage{color,soul}%
\usepackage[para]{threeparttable}%
\usetikzlibrary{matrix, positioning}%
\usepackage[normalem]{ulem}%
\usepackage{hyperref}%
\usepackage{rotating}%
\usepackage{subcaption}%
\usepackage{nicematrix}%
\usetikzlibrary{arrows.meta}
\definecolor{gray1}{RGB}{0.5, 0.5, 0.5}
\definecolor{red1}{RGB}{255,188,217}
\definecolor{green}{RGB}{204,255,204}
\definecolor{blue1}{RGB}{204,229,255}
\definecolor{purple1}{RGB}{229,204,255}
\definecolor{orange1}{RGB}{255,229,204}
\definecolor{white}{RGB}{255,255,255}
\hypersetup{
    colorlinks=true,
    linkcolor=blue,
    filecolor=black,      
    urlcolor=pink,
    citecolor=black,
}
\AtEndEnvironment{threeparttable}{\vskip10pt}{}{}
\newtcbox{\badge}[1][red1]{
  on line, arc=2pt, colback=#1,
  colframe=#1, fontupper=\bfseries\color{black},
  boxrule=1pt, boxsep=0pt,
  left=1pt, right=1pt, top=1pt, bottom=1pt
}
\usepackage{tikz}
\newcolumntype{C}[1]{>{\centering\arraybackslash}p{#1}}

\definecolor{azure(colorwheel)}{rgb}{0.0, 0.5, 1.0}

\begin{document}

\begin{frontmatter}

\title{Deep learning for precipitation nowcasting: \\A survey from the perspective of time series forecasting}

\author[label1]{Sojung An}\ead{sojungan@kiaps.org}
\author[label1]{Tae-Jin Oh}\ead{oht@kiaps.org}
\author[label2]{Eunha Sohn}\ead{soneh0431@korea.kr}
\author[label3]{Donghyun Kim\corref{cor1}}\ead{d\_kim@korea.ac.kr}
\cortext[cor1]{Corresponding author.}
\affiliation[label1]{organization={Korea Institute of Atmospheric Prediction Systems},
            addressline={35, Boramae-ro, Dongjak-gu}, 
            city={Seoul},
            postcode={07071}, 
            country={Republic of Korea}}
\affiliation[label2]{organization={National Meteorological Satellite Center},
            addressline={61-18, Guam-gil, Gwanghyewon-myeon}, 
            city={Jincheon-gun},
            postcode={27803}, 
            country={Republic of Korea}}
\affiliation[label3]{organization={Korea University},
            addressline={145, Anam-ro, Seongbuk-gu}, 
            city={Seoul},
            postcode={02841}, 
            country={Republic of Korea}}

\begin{abstract}
Deep learning-based time series forecasting has dominated the short-term precipitation forecasting field with the help of its ability to estimate motion flow in high-resolution datasets.
The growing interest in precipitation nowcasting offers substantial opportunities for the advancement of current forecasting technologies.
Nevertheless, there has been a scarcity of in-depth surveys of time series precipitation forecasting using deep learning.
Thus, this paper systemically reviews recent progress in time series precipitation forecasting models.
Specifically, we investigate the following key points within background components, covering: i) preprocessing, ii) objective functions, and iii) evaluation metrics.
We then categorize forecasting models into \textit{recursive} and \textit{multiple} strategies based on their approaches to predict future frames, investigate the impacts of models using the strategies, and performance assessments. 
Finally, we evaluate current deep learning-based models for precipitation forecasting on a public benchmark, discuss their limitations and challenges, and present some promising research directions.
Our contribution lies in providing insights for a better understanding of time series precipitation forecasting and in aiding the development of robust AI solutions for the future.
\end{abstract}

\end{frontmatter}

\section{Introduction}
\label{sec:introduction} 
Over the past decades, the world has relied on weather forecasts to warn the public about hazardous weather, agriculture, and energy use. 
Accurate precipitation nowcasting, in particular, has gained more attention for preventing severe weather-related damage due to the high degree of uncertainty in extreme precipitation cases.
The concept of precipitation nowcasting emerged with the advancement of radar sensor \citep{browning1989}.
\begin{tcolorbox}[colback=yellow!30!gray!15, colframe=yellow!30!black!50]
\textbf{Precipitation Nowcasting} is the short-term precipitation forecasting (up to 6 hours) of providing highly detailed predictions of localized weather phenomena that undergo significant changes.
\end{tcolorbox}\noindent
Traditional nowcasting models primarily relied on extrapolating radar observations with the Lagrangian advection assumption \citep{germann2002}.
The systems limit their ability to forecast the formation or dissipation of rainfall, resulting in predictions primarily focused on rainfall movement. 
With the increasing demand for accurate precipitation forecasting to mitigate potential damages, the research scope has expanded to apply machine learning \citep{shi2015,wang2017}.
This expansion aligns with the broader trend seen in recent decades, where deep learning (DL) has increasingly been utilized across various domains, with precipitation nowcasting being no exception.

DL-based techniques for time series forecasting have acted as a significant milestone in precipitation forecasting.
For the problem of precipitation nowcasting, \cite{shi2015} formulated precipitation nowcasting from radar observations as a time series forecasting task.
They proposed ConvLSTM which estimates motion flow from high-resolution datasets, thereby enhancing the prediction accuracy of extreme weather events.
Subsequent studies have leveraged DL for precipitation nowcasting, achieving better performance compared to the traditional methods.
For example, DGMR \citep{ravuri2021}, a 90-minute (min) prediction system, demonstrated its potential in time series forecasting.
However, radar-based nowcasting models face scalability limitations due to the fixed locations of radar systems.
Moreover, the installation of radar systems is expensive and requires agreements with local authorities and communities, as well as trained personnel for operation.
From this perspective, combining satellite imagery with radar data has surfaced as a promising approach to overcome the constraints of radar-based nowcasting \citep{lebedev2019, sonderby2020, horvath2021}.
Although satellites primarily observe clouds rather than precipitation directly, satellite imagery exhibits features that significantly influence precipitation. 
By leveraging satellite data, it is possible to address these limitations and expand the scope of nowcasting to cover larger territories.

With increasing datasets becoming publicly available, researchers are exploring innovative approaches to predict precipitation from the point of view of time series forecasting.
While precipitation forecasting has attracted the attention of the research community, few technical surveys have been conducted in this area.
Indeed, it is still challenging to find datasets for precipitation nowcasting and to stay aware of the latest developments.
This requires a review on concepts involving process, dataset information, evaluation, and many others that are relevant to the nowcasting and its strategy of designing DL models.
Through the extensive examination conducted in our survey, we advocate fostering faster and more comprehensive progress in exploring the domain of precipitation forecasting.
The main contributions of this survey can be summarized in five-fold.

\begin{enumerate}
  \item[\textbf{(1)}] \textbf{Analyze key challenges caused by real-world datasets and provide practical implications for preprocessing observation data.} Observation data (e.g., radar reflectivity and satellites) have been less investigated than commonly used images. This limited exploration hinders broader application and advancement of knowledge in this area, thus confining the domain of precipitation forecasting to specific research groups. To improve the capability of precipitation forecasting, we made greater efforts to bridge the knowledge between data preprocessing (e.g., normalization and sampling) and different sensor datasets.
  \item[\textbf{(2)}] \textbf{Outline effective objective functions in precipitation forecasting and provide basic evaluation criteria.} As numerous research works have been published on precipitation forecasting models, we analyze approaches to precipitation forecasting and review them depending on how previous literature design objective functions for scientific modeling. We then provide an overview to generalize the methods of reliability assessment of precipitation forecasting.
  \item[\textbf{(3)}] \textbf{Present a comprehensive review of relevant works categorized based on operational principles, summarizing the current state-of-the-art.} In comparison with other related reviews, our work surveys up-to-date technology of applications and is not restricted to a few network frameworks. In the context of architecture, this study reviews various applications, encompassing Diffusion- and Transformer-based models for precipitation forecasting.
  \item[\textbf{(4)}] \textbf{Compare the performance of precipitation forecasting models and offers insights into their methodologies.} We conduct a comparison of prominent models to assist researchers interested in extending their capabilities to real-life systems. Moreover, we evaluate the effectiveness of the models across various thresholds and over time, highlighting unresolved challenges within this field. To the best of our knowledge, this is the first review paper to present comparison results for DL models using benchmark datasets.
  \item[\textbf{(5)}] \textbf{Engage in a discussion of various research directions and outline prevailing challenges in the field.} While many researchers are making progress in precipitation forecasting, there is still room for further advancements. Precipitation forecasting models simulate the dynamics of natural phenomena, and learning representations from sparse precipitation data degrades the performance of AI models. Additionally, predicting changes in long-term trends or fusing the multi-sensor dataset remains an unsolved problem. Therefore we discuss potential solutions to address these problems, exploring in a discussion of various research directions.
\end{enumerate}

The structure of the paper is as follows: Section~\ref{sec:bakckground} presents the background on time series precipitation forecasting and compares related surveys.
Section~\ref{sec:dataset} provides a prominent dataset in precipitation nowcasting and Section~\ref{sec:preprocessing} reviews preprocessing techniques for observation data from different sensors.
Section~\ref{sec:object} introduces the proposed objective function for precipitation forecasting and Section~\ref{sec:evaluation} gives an overview of the basic evaluation functions.
Next, Section~\ref{sec:recursive_strategy} and Section~\ref{sec:multiple_strategy} present a comprehensive review that categorizes existing literature on DL-based models. Then, Section~\ref{sec:comarison_evaluation} provides the comparison and evaluation on a public benchmark, and Section~\ref{sec:future} discusses future research directions.
Finally, the conclusions are presented in Section~\ref{sec:conclusion}.

\section{Background and related works}
\label{sec:bakckground}
This section introduces preliminaries relevant to this survey, including the taxonomy of precipitation nowcasting (Section \ref{sec:preliminaries}), and provides an analysis of the existing surveys in deep learning applied to nowcasting (Section \ref{sec:related_works}).

\begin{figure*}[t]
    \centering
    \begin{tikzpicture}[
        level 1/.style = {black, sibling distance = 5.5cm, level distance = 2.5cm, text=black},
        level 2/.style = {black, sibling distance = 2.2cm, level distance = 1.5cm, text=black},
        edge from parent fork down
        visible on/.style={draw=none,fill=none,text=black},
        invisible/.style={visible on=,draw=none},
    ]
    \normalsize
    \node [align=center] {\textbf{  Precipitation nowcasting problem}\\
        \begin{tabular}{ccccccc}
            $x_{1}$ & \multirow{2}{*}{$\cdots$} & $x_{m}$ & \multirow{2}{*}{$\rightarrow$} & $x_{m+1}$ & \multirow{2}{*}{$\cdots$} & $x_{m+n}$ \\
            \fbox{\includegraphics[width=.08\textwidth]{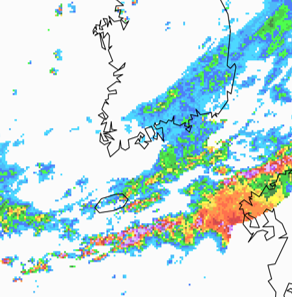}} & & \fbox{\includegraphics[width=.08\textwidth]{fig_x1.png}} & & \fbox{\includegraphics[width=.08\textwidth]{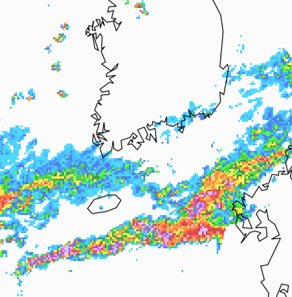}} &  & \fbox{\includegraphics[width=.08\textwidth]{fig_x2.png}}\\
        \end{tabular}}
        child {node[align=center]{\textbf{Recursive strategy}\\\includegraphics[width=.15\textwidth]{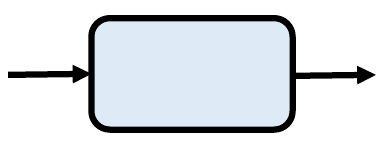}}
            child {node[text width=3.5cm] {\scriptsize{\badge{N}} \normalsize{Non-adversarial}}}
            child {node {\scriptsize{\badge[blue1]{A}}\normalsize{Adversarial}}}
            edge from parent [dashed]}
        child {node[align=center]{\textbf{Multiple strategy}\\\includegraphics[width=.15\textwidth]{fig_box.pdf}}
            child {node[text width=1.2cm] {\scriptsize{\badge[green]{U}} \normalsize{UNet}}}
            child {node[text width=1.8cm] {\scriptsize{\badge[purple1]{D}} \normalsize{Diffusion}}}
            child {node {\scriptsize{\badge[orange1]{T}} \normalsize{Transformer}}}};
    \node at (-4.4, -2.65) {$x_{1:m}$}; 
    \node (ht) at (-2.75, -2.65) {$h_t$}; 
    \node at (-0.8, -2.65) {$x_{m+1:m+n}$};
    \node at (1.1, -2.65) {$x_{1:m}$}; 
    \node at (2.75, -2.65) {$h$}; 
    \node at (4.7, -2.65) {$x_{m+1:m+n}$};
    \draw[thick, black, -{Triangle}] (ht) ++(0.6, -0.15) +(0:0.15cm) arc[start angle=40, end angle=-220, x radius=1cm, y radius=0.3cm];
    \end{tikzpicture} 
    \caption{Paradigms in precipitation nowcasting. We classify forecasting models at the first level based on two training strategies: the recursive strategy and the multiple strategy. While the recursive strategy predicts future time steps sequentially at each step $t$, the multiple strategy predicts future frames simultaneously. At the second sub-level, models of the recursive strategy are classified into non-adversarial and adversarial-based categories, and multiple strategy models are categorized into UNet, Diffusion, and Transformer.}
    \label{fig:tree}
\end{figure*}

\subsection{Preliminaries of time series precipitation forecasting}
\label{sec:preliminaries}
Time series precipitation forecasting aims to predict future time steps using historical observations. Given historical observation data $\mathbf{X}=\left\{x_{1}, x_{2}, \cdots, x_{m}\right\} \in \mathbb{R}^{c\times m\times h \times w}$, we denote the number of variates, the input frames, height, and width as $c, m, h$, and $w$, respectively. Time series forecasting aims to predict $\mathbf{Y}=\left\{x_{m+1}, x_{m+2}, \cdots, x_{m+n}\right\}\in\mathbb{R}^{n \times h \times w}$, where $n$ refers to future time frames. There are two different approaches to predicting future time frames in time series data: univariate input data ($c$=1), which involves a series with a single time-dependent radar dataset \citep{ravuri2021,zhang2023}, and multivariate input data ($c>$1), which incorporates satellite datasets or additional weather variables \citep{sonderby2020,fernandez2021}. Note that we define the forecasting problem as the same for both types of data. The ground-truth pairs are set to \textbf{GT} = $\bigl\{x_{1:m}, \,x_{m+1:m+n} \bigl\}$.
Let the forecasting model $G_{\theta}$ with parameters $\theta$ have the objective function (denoted $\mathcal{L}(\theta)$) that minimizes future ground-truth time series observations and prediction results, as follows:
\begin{equation}
    \mathcal{L}(\theta) = \mathbb{E} (G_{\theta}(h; x_{1:m} - x_{m+1:m+n}),
\end{equation}
where $h$ represents hidden states. Many DL models widely used in computer vision tasks have been proposed to boost the performance of nowcasting, such as convolutional neural networks \citep{sonderby2020, zhang2023}, recurrent neural networks \citep{wang2022, she2023}, Transformers \citep{gao2022,ning2023}, and so on. Recently, with the boom of generating radar frames with realistic radar frames, generative models have attracted much attention various models came out, such as adversarial-based models, e.g., DGMR \citep{ravuri2021} and diffusion-based models, e.g., Prediff \citep{gao2023}.

Figure \ref{fig:tree} summarizes the precipitation nowcasting approaches categorized by two strategies and some relevant subsets of the categories. The recursive strategy recursively predicts future time steps one at a time and updates the forecast at each time step. Given a sample time step $t \in \{ m+1, \cdots, m+n \}$, a model $G_\theta$ involves to predict the value at the next time step $x_{t+1}=G_{\theta}(h_t; x_{1:t})$, where $h_t$ represents hidden state at time step $t$.
To be more specific, the recursive learning problem can be represented as:
\begin{equation}
    P(Y|X)=\prod_{t=m+1}^{m+n}P(x_{t}|x_{1}, \cdots, x_{t-1}).
\end{equation}
This approach updates the hidden states at each time step and repeats the process until the final time step is reached. In the multiple strategy, the future time steps are predicted with a hidden state $h$ as follows: $x_{\forall t }=G_{\theta}(h; x_{1:m})$. The approach aims to forecast multiple time steps ahead at once, and can be expressed as follows:
\begin{equation}
    P(Y|X)=P(x_{m+1}, \cdots,  x_{m+n} | x_{1}, \cdots, x_{m}).
\end{equation}
This means that models learn the time dependence based on a hidden state to estimate $n$ future steps with respect to $m$ time steps, not each individual time step.

\subsection{Related surveys}
\label{sec:related_works}
%
In recent years, several surveys surfaced on the topic of precipitation nowcasting to explore the methods that have been used in this field and to identify areas for future improvement in performance.
Existing surveys on precipitation nowcasting can be categorized into two groups. 
\begin{enumerate}
    \item[\textbf{(1)}] \textbf{High-level overview:} Most surveys emphasize the ongoing challenge of balancing model accuracy with computational efficiency in weather nowcasting, particularly regarding the limitations of current approaches in predicting extreme weather events \citep{ashok2022, verma2023, upadhyay2024, salcedo2024}. However, these papers provide a comprehensive overview rather than a technical comparison of DL-based precipitation forecasting, arranged with statistical methods and numerical weather prediction (NWP)-based approaches. Therefore, they omit methodological discussions regarding strengths and weaknesses, leading to a limited understanding of practical constraints when constructing DL models.
    \item[\textbf{(2)}] \textbf{Methodological analysis:} These surveys categorized nowcasting models into stochastic and deterministic methods, aiming to enhance understanding of the research field by reviewing DL-based models from a technical standpoint \citep{prudden2020, gao2021}. The studies highlight key research areas with pointers to the relevant sections, but the surveys are previous methods. The absence of recent architectures in existing literature restricts the comprehensive analysis of state-of-the-art nowcasting models. This limitation requires the need for a survey paper to fill the gap and provide a more inclusive examination of recent nowcasting methodologies.
\end{enumerate}

After conducting an analysis of existing survey papers, we identified a gap in the literature - a lack of an in-depth survey on precipitation nowcasting that investigates their performances across recent methodological approaches.
Not only does our research aim to provide a more theoretically and practically grounded understanding of precipitation nowcasting, but it also offers a systematic review of the overall process for initiating nowcasting research.
Note that the majority of the work presented is related to modeling for sensor datasets and that many varieties of DL frameworks are often based on radar-based nowcasting. 
This survey does not differentiate between univariate forecasting using radar sensors and multivariate forecasting by combining other sensors, and review nowcasting applications.

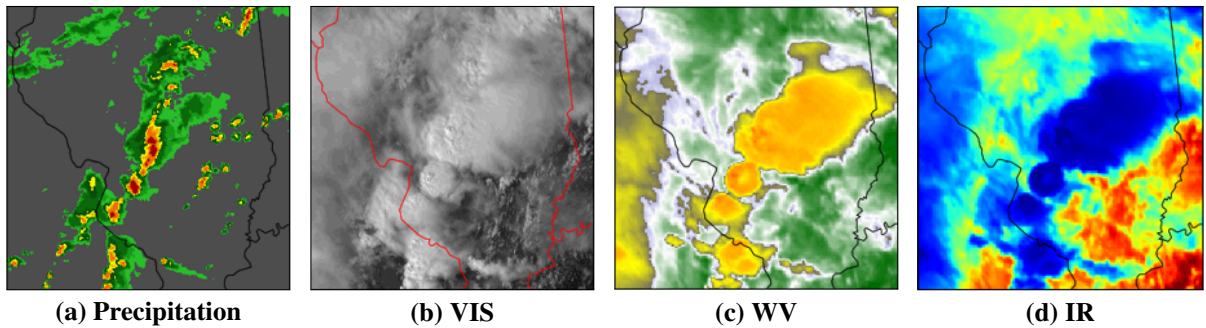
\begin{figure*}[tb]
    \centering
    \begin{tikzpicture}
    \node at (-6,-2.2) {\textbf{(a) Precipitation}};
    \node at (-2,-2.2) {\textbf{(b) VIS}};
    \node at (2,-2.2) {\textbf{(c) WV}};
    \node at (6,-2.2) {\textbf{(d) IR}};
    \node at (-6,0) {\fbox{\includegraphics{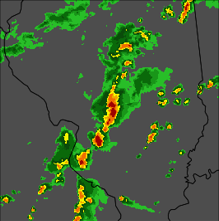}}};
    \node at (-2,0) {\fbox{\includegraphics{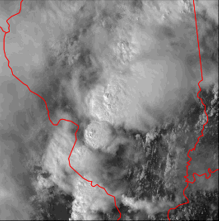}}};
    \node at (2,0) {\fbox{\includegraphics{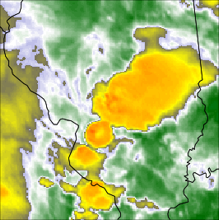}}};
    \node at (6,0) {\fbox{\includegraphics{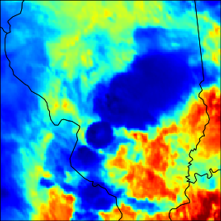}}};
    \end{tikzpicture}
    \caption{Visualization of the SEVIR dataset from April 30, 2019, at 18 UTC. The images depicting weather events captured over the contiguous US. (a) Vertically integrated liquid of NEXRAD radar. (b) GOES-16 satellite channel 2 visible (VIS). (c) GOES-16 satellite channel 9 water vapor (WV). (d) GOES-16 satellite channel 13 infrared (IR). (Source) The images were obtained from the SEVIR official page.}
    \label{fig:sample}
\end{figure*}

 \section{Datasets for precipitation nowcasting}
\label{sec:dataset}
This section discusses the properties of the sensory dataset used for precipitation nowcasting.
We specifically explore the data processing for weather radar (Section~\ref{subsec:weather_radar}) and for the weather satellites (Section~\ref{subsec:weather_satellite}). Then, we highlight how the two different sensory datasets are combined (Section~\ref{subsec:fusion}), and a comprehensive review of the available datasets for each sensor type is presented (Section~\ref{subsec:dataset}). Figure \ref{fig:sample} shows an example of an RGB composite image in the SEVIR dataset \citep{veillette2020}, which is visualized across four different data types: (a) composite radar imagery, (b) visible satellite imagery, (c) Water vapor satellite imagery, and (d) Infrared satellite imagery

\begin{table*}[tb]
    \centering
    \begin{tabular}{C{1cm}*{8}{C{1.45cm}}}
    \Xhline{2\arrayrulewidth}
    & \scriptsize{\textbf{HKO-7}} & \scriptsize{\textbf{IowaRain}} & \scriptsize{\textbf{RYDL}} & \scriptsize{\textbf{MRMS}} & \scriptsize{\textbf{Shanghai}} & \scriptsize{\textbf{SEVIR}} & \scriptsize{\textbf{OPERA}} & \scriptsize{\textbf{Meteonet}}\\
    & \tiny{\citep{shi2017}} & \tiny{\citep{sit2021}} & \tiny{\citep{ayzel2020}} & \tiny{\citep{mrms}}
    & \tiny{\citep{chen2020}} & \tiny{\citep{veillette2020}} & \tiny{\citep{herruzo2021}} & \tiny{\citep{meteonet2020}} \\
    \hline
    \scriptsize{Year} & \scriptsize{09--15} & \scriptsize{16--19} & \scriptsize{14--15} & \scriptsize{17--17} & \scriptsize{15--18} & \scriptsize{17--19} & \scriptsize{19--21} & \scriptsize{16--18}\\ 
    \scriptsize{Frequency} & \scriptsize{6 min} & \scriptsize{5 min} & \scriptsize{5 min} & \scriptsize{2 min} & \scriptsize{6 min} & \scriptsize{5 min} & \scriptsize{15 min} & \scriptsize{5--15 min} \\
    \scriptsize{Resolution} & \scriptsize{2 km} & \scriptsize{0.5 km} & \scriptsize{1 km} & \scriptsize{1 km} & \scriptsize{1 km} & \scriptsize{0.5--8 km} & \scriptsize{2 km} & \scriptsize{1 km}\\
    \scriptsize{Coverage} & \scriptsize{Hong Kong} & \scriptsize{US} & \scriptsize{Germany} & \scriptsize{US}  & \scriptsize{Shanghai} & \scriptsize{US} & \scriptsize{Europe} & \scriptsize{France} \\
    \scriptsize{Format} & \scriptsize{png} & \scriptsize{binary} & \scriptsize{hdf5} & \scriptsize{grib2} & \scriptsize{binary} & \scriptsize{hdf5} & \scriptsize{hdf5} & \scriptsize{hdf5}\\
    \scriptsize{Type} & \scriptsize{\textcolor{teal}{\textcircled{R}}} & \scriptsize{\textcolor{teal}{\textcircled{R}}} & \scriptsize{\textcolor{teal}{\textcircled{R}}} & \scriptsize{\textcolor{teal}{\textcircled{R}}} & \scriptsize{\textcolor{teal}{\textcircled{R}}} & \scriptsize{\textcolor{teal}{\textcircled{R}}\textcolor{brown}{\textcircled{S}}} & \scriptsize{\textcolor{teal}{\textcircled{R}}\textcolor{brown}{\textcircled{S}}} & \scriptsize{\textcolor{teal}{\textcircled{R}}\textcolor{brown}{\textcircled{S}}} \\
    \scriptsize{URL} & \scriptsize{\href{https://github.com/sxjscience/HKO-7}{[link]}} & \scriptsize{\href{https://github.com/uihilab/IowaRain}{[link]}} & \scriptsize{\href{https://zenodo.org/records/3629951}{[link]}} & \scriptsize{\href{https://data.eol.ucar.edu/dataset/541.033}{[link]}} & \scriptsize{\href{https://dataverse.harvard.edu/dataset.xhtml?persistentId=doi:10.7910/DVN/2GKMQJ}{[link]}} & \scriptsize{\href{https://registry.opendata.aws/sevir/}{[link]}} & \scriptsize{\href{https://github.com/agruca-polsl/weather4cast-2023}{[link]}} & \scriptsize{\href{https://meteonet.umr-cnrm.fr}{[link]}}\\
     \Xhline{2\arrayrulewidth}
    \end{tabular}
    \caption{Summary of publicly available sensor dataset resources. \textbf{Year}: data collected years (represented as the last two digits of the year), \textbf{Frequency}: time-frequency, \textbf{Resolution}: spatial resolution of data, \textbf{Coverage}: observed coverage, \textbf{Format}: data file format, \textbf{Type}: Radar \textcolor{teal}{\textcircled{R}}, Satellite \textcolor{brown}{\textcircled{S}}, \textbf{URL}: data URL link. \textcolor{blue}{[link]} directs to dataset websites.}
    \label{tab:data}
\end{table*}

\subsection{Weather radar}
\label{subsec:weather_radar}
Weather radar sends pulses into the air to detect rainfall and estimate its type (e.g., rain, snow, etc).
The radar sensors can provide reflectivity data consisting of one channel in decibels (dBZ).
Rainfall intensity can be categorized based on the radar reflectivity values; $[-35, 20)$: small hydrometeors, $[20, 35)$: light rain, $[35, 50)$: medium rain, and $[50, \infty]$: heavy precipitation \citep{binetti2022}. 
Small hydrometeors are useful for detecting very dry light snow or drizzle that has lower reflectivity.
Typically, the \textit{Z-R relationship} is a step in radar-based quantitative precipitation estimation that involves converting reflectivity values into rainfall intensity. 
The reflectivity is converted by the \textit{Z-R relationship} between the radar reflectivity factor $Z\ (\text{mm}^6\ \text{m}^{-3})$ and rain rate $R\ (\text{mm}\ \text{h}^{-1})$ as follows: $Z = aR^b,$ where $a$ and $b$ are parameters obtained empirically depending on the precipitation type.
A fixed \textit{Z-R relationship}, such as $a=200$ and $b=1.6$ \citep{marshall1948} can be used; however, \cite{Kim2021aa} indicates that calibrating the relationship variability via Bayesian regression for each radar-gauge pair can produce a better fit.

\subsection{Weather satellite}
\label{subsec:weather_satellite}
Weather satellites monitor the weather and climate of the Earth and detect the movement of storm systems and other cloud patterns.
Currently, there are more than 4000\footnote{\scriptsize{
\url{https://www.ucsusa.org/resources/satellite-database}}} satellites orbiting the Earth, each with its unique data-gathering sensor.
Geostationary satellites used in time series forecasting are equipped with calibrated sensors to detect two-dimensional channels, including infrared, visible, and water vapor.
It is important to note that visible is only available during the daytime, as the data represent solar radiation in albedo. \textit{Why is satellite data, which detects cloud information, important in precipitation forecasting?} Satellites highlight specific phenomena provide signals related to precipitation and provide global coverage that complements radar observations. Metnet-v2 \citep{espeholt2022} found that satellites play an important role not only in precipitation nowcasting but also in longer predictions, contributing to atmospheric moisture corrections and improving cloud detection. \cite{Lee2021} used high-resolution satellite imagery from GEOS-16 with brightness temperature to detect convection. Global data were available at 1--2 km resolution at 10--15 min intervals, whereas higher resolution local data were available at 500 m resolution at 2 min intervals. Satellite-based precipitation products can complement the radar data by providing coverage over larger geographical areas. This can involve adjusting the rainfall features data based on the various angles of elevation of satellite \citep{niu2024}, or using the satellite data to fill in gaps where radar coverage is limited \citep{an2023}.

\subsection{Sensor fusion}\label{subsec:fusion}
Sensor fusion is a crucial technique for achieving accurate precipitation nowcasting by combining observations from ground-based radars and satellites. Radars and satellites often provide observations at different times and frequencies depending on data resolution. In addition, fusing high-resolution radar and satellite data over large areas requires significant computational resources. The key challenge in sensor fusion for nowcasting is aligning the different spatial and temporal characteristics of radar and satellite observations to produce a coherent, high-resolution precipitation forecast. Here, we highlight the key points about data fusion for precipitation nowcasting for the same \citep{sonderby2020,seo2022} or different \citep{an2023} coverage areas. When both sets of observational data cover the same area, images are merged by concatenating them along the channel axis. Assuming there are $d$ channels in the satellite data, the input size would be $X \in \mathbb{R}^{(d+1) \times h \times w}$ \citep{sonderby2020,seo2022}. For a longer forecasting lead time, \cite{an2023} merged satellite data, which is four times larger than the radar dataset, by tokenizing the satellite data to align with the radar dimensions. The model incorporates positional information via periodic functions to enhance the data fusion process, resulting in a combined input of $X \in \mathbb{R}^{5 \times h \times w}$. Considering cloudiness as an additional layer of information helps investigate the influence of different cloud and atmospheric parameters \citep{andrychowicz2023}. This approach also allows us to assess the likelihood of precipitation persistence in these regions and to identify the interactions of additional external climate forcings like radiation \citep{samset2016}. 

\subsection{Available datasets}\label{subsec:dataset}
Table \ref{tab:data} summarizes the radar and satellite data used in previous studies. We have the following data type summaries for precipitation forecasting: \textcircled{1} radar (HKO-7, IowaRain, RYDL, MRMS, and Shanghai) and \textcircled{2} radar \& satellite (SEVIR, OPERA, and MeteoNet).
Publicly available data typically consists of high-resolution images captured at short intervals (5 to 6 min) with diverse data formats. The benchmark dataset was collected from the original data sources by minimizing data bias and noise. The benchmark datasets were conducted in data selection processes \citep{shi2015, chen2020, veillette2020, sit2021} based on reflectivity intensity. For instance, only rain days were chosen for the dataset \citep{shi2015}, or those with at least 0.5 mm precipitation over 10\% of the domain \citep{sit2021}. Careful data selection helps mitigate selection alleviate the bias of the dataset and enhance generalize to the broader population. Noise reduction techniques, such as removing pixels with low correlation coefficients or applying consecutive snapshot requirements, were implemented to improve data quality \citep{chen2019, ayzel2020, sit2021}. In the case of the Shanghai dataset \citep{chen2020}, the authors removed pixels with correlation coefficients of less than 0.85 \citep{kumjian2010} to eliminate anomalies. In terms of data types, the gathered data can be in binary or PNG data formats. Datasets with PNG format are visual representations of this data which can provide valuable insights and help improve the interpretation of the radar data; however, the dataset is hard to use as scientific ground-truth. Non-uniform color mapping in radar-derived images can pose challenges for precise precipitation prediction, as it may affect the interpretation of intensity levels. There is a lack of widely accepted standards for evaluating radar reflectivity datasets.

\section{Preprocessing}
\label{sec:preprocessing}
Real-world datasets often do not follow a normal distribution, and imbalanced data with noise can cause performance issues. When some class is sparsely distributed compared to others, the model tends to overfit the majority class during training, resulting in poor performance of the minority class. Preprocessing is important for the efficient and accurate comparison of AI models using noisy real-world data, and this process contributes to improving the performance of the models. This section presents the effective techniques for precipitation data: \textcircled{1}  clipping (Section \ref{subsec:clipping}), \textcircled{2} scaling (Section \ref{subsec:scaling}), \textcircled{3}  sampling (Section \ref{subsec:sampling}), and \textcircled{4} sliding window (Section \ref{subsec:sliding}). To refine the radar dataset in constructing datasets, several preprocessing steps, as delineated in Table \ref{tab:radar}, are commonly applied.

\subsection{Clipping}\label{subsec:clipping}
The dataset undergoes clipping to maximum rainfall or reflectivity, which addresses the sparsity of rainfall data and imbalance issues. Real-world datasets often exhibit high-class imbalance, with the majority of instances concentrated near zero. A study in Beijing \citep{ma2019} found that raindrops of diameter 0.9–2.5 mm contributed most to accumulated rainfall. In other regions, precipitation exceeding 10 \text{mm} is scarce, accounting for less than 1\% \citep{choi2022, kim2022}. These imbalanced datasets are problematic, as they often contain much higher error rates in the tail of the distribution. Researchers often set the maximum rainfall or reflectivity values to 100 to 128 \citep{ravuri2021, zhang2023} or 70 to 76 \citep{meteonet2020, yu2023}, respectively. This approach enables the handling of sparse datasets, spanning from the maximum output setting to the highest rainfall levels. When the dataset with high variance, occurs when a model fits too closely to the training data and is then unable to generalize to unseen dataset \citep{ravuri2021}.

\begin{table}[bt]
    \centering
    \begin{threeparttable}
    \renewcommand{\arraystretch}{1.2}{
    \begin{tabular}{>{\centering\arraybackslash}m{2.2cm}|>{\centering\arraybackslash}m{5.8cm}}
    \hline
    \textbf{Components} & \textbf{Methods}\\
    \hline
    \small{Clipping} & \small{100 mm\tnote{a} , 128 mm\tnote{b} , 70 dBZ\tnote{c} , 76 dBZ\tnote{d}}\\
    \small{Scaler} & \small{$z$-score}\tnote{e} , \small{minmax}\tnote{f} , \small{logarithmic}\tnote{g} \\
    \multirow{2}{*}{\small{Sampling}} & \small{\textcircled{1} $\mathcal{P}r^1(X)=\frac{1}{(m+n)hw}\sum_{\forall x'_{t} \subset X'}x'_t$}\\
    & \small{\textcircled{2} $\mathcal{P}r^2(X) =\sum (1-exp^{-x_t})+\epsilon$} \\
    \small{Sliding window} & \small{20 min\tnote{h} , 60 min\tnote{i} , 120 min\tnote{j}} \\
    \hline
    \end{tabular}}
    \begin{tablenotes}
    \scriptsize{
    $\bullet$ Clip:
    \item[a] \cite{an2023} 
    \item[b] \cite{ravuri2021,zhang2023}
    \item[c] \cite{meteonet2020}
    \item[d] \cite{yu2023} \\
    $\bullet$ Scaler:
    \item[e] \cite{jeong2021,bakkay2022}
    \item[f] \cite{shi2015,wang2017} \\ 
    \item[g] \cite{li2021, espeholt2022, geng2023}\\
    $\bullet$ Sampling: \\ 
    \textcircled{1} $\mathcal{P}r_{min}^{1}$=3$\times 10^{-2}$ \cite{an2023}, $\mathcal{P}r_{min}^{1}$=1$\times 10^{-1}$ \cite{ayzel2020}\\
    \qquad \textcircled{2} $\mathcal{P}r_{min}^{2}$=2$\times 10^{-4}$ \cite{ravuri2021} \cite{zhang2023}\tnote{\#}}\\
    $\bullet$ Sliding: 
    \item[h] \cite{fernandez2021, zhang2023, leinonen2023}\\
    \item[i] \cite{fernandez2021, reulen2024, gao2022}\\
    \item[j] \cite{shi2015,meteonet2020}
    \end{tablenotes}
    \end{threeparttable}
    \vspace{-1em}
    \caption{Brief details of preprocessing approaches in the literature. \textbf{Clip}: clipping value to handle imbalanced data setting to maximum rainfall. \textbf{Scaler}: data normalization methods (minmax: min-max scaler, log: logarithmic scaler). \textbf{Sampling}: sampling methods to prevent overfitting in sparse datasets. Zhang$^\text{\#}$ conducted hierarchical sampling in training, which involves first sampling from the full image and then sampling the cropped. \textbf{Sliding}: sliding window.\\
    \textbf{Notation.} Let the $X'$ be a conditional matrix of input frames $X$, where the value is 1 if the value exists and 0 otherwise. Given the probability score of sampling as $\mathcal{P}r(X)$, the training dataset is sampled to sequences as $\mathcal{P}r(X) \geq \mathcal{P}r_{min}$, where $\mathcal{P}r_{min}$ is a minimum probability for controlling the overall inclusion rate. Here, $x_t$ represents an input frame, and $\epsilon$ is a small constant.}
    \label{tab:radar}
\end{table}

\subsection{Scaling}\label{subsec:scaling}
Data scaling can improve the performance of DL models by making the learning process stable, especially on time series forecasting tasks where the data is highly non-linear and dynamic. The scaling techniques for data normalization put all features on a similar scale, allowing the model to learn from a more balanced distribution and making the optimization process more robust for DL models \citep{huang2023}. Three commonly used techniques for nowcasting include: $z$-score normalization, logarithmic transformation, and min-max normalization. To train satellite and radar sensor data with different ranges, it is common to normalize variables by applying $z$-score to them. Sensory datasets have a heavy-tailed distribution, as the sensory images contain values that can have extreme outliers or values in the tail ends of the distribution \citep{achim2003}. To handle satellite datasets, it is better to use percentile-based normalization techniques, like $z$-score normalization, rather than simple min-max normalization \citep{herruzo2021}. These existing scaling methods are just done to ensure that features with different scales don't unduly influence the training process \citep{jeong2021, li2023}, while the ML models may still encounter challenges in adequately handling outliers.

\subsection{Sampling}\label{subsec:sampling}
Sampling methods are another way to address the data imbalance problem, encompassing both over-sampling and under-sampling approaches. These techniques modify the training distribution in order to decrease the level of imbalance or alleviate data noise by anomalies. In their simplest forms, the dataset is applied to over-sampling with rainy cases \citep{shi2015}. Or, recent studies have presented that sample by grouping a distribution bin or adjusting the proportion of rainy pixels \citep{ayzel2020, zhang2023}. For example, DGMR \citep{ravuri2021} proposed a sampling strategy using acceptance score. They remove samples from the over-represented class by using a defined sampling probability. Given the probability of $i$-th element of a sample data $x$ as $q_{x_i}=\sum 1-exp(-x_i)$, samples above the $q_{x_i}$ are included in the training set. Fitting data distribution into equal-width bins is one of the keys to robust model performance in diverse precipitation cases.

\subsection{Sliding window}\label{subsec:sliding}
The sliding window approach is a widely used term in time series forecasting tasks.  The term refers to a technique where a fixed-size window of consecutive time steps is used as input to predict the next time step or a sequence of future time steps.  The window size is an important hyperparameter that needs to be tuned in the forecasting task, which is typically set between 20 min \citep{ravuri2021} to 2 hours (h) \citep{meteonet2020} in nowcasting models. Building on previous studies, DL models can effectively capture temporal dependency and patterns within precipitation data by adopting the sliding window with intervals exceeding 20 min and considering a fixed number of previous time steps as input to forecast future rainfall intensity.

\section{Objective functions}
\label{sec:object}
The effectiveness of deep learning models is largely dependent on their objective functions, which are crucial for precipitation forecasting. The objective function is a decisive component for precipitation nowcasting, as it shapes the learning process and determines the quality of the forecasts in terms of accuracy, and the ability to capture important high-intensity precipitation events. In this context, certain challenges, such as data imbalance, blurriness, and data sparsity, need to be taken into account.

\subsection{Weighted loss}
The weighted loss function assigns weights to the learning objective function based on precipitation intensity.
In precipitation forecasting, where data imbalance is addressed, applying weights can prioritize high-intensity rainfall.
This means that errors made on predictions of heavy rainfall will contribute more to the overall loss function than errors made on predictions of light rainfall.
In TrajGRU \citep{shi2017}, they implemented a strategy of dividing rainfall intensity into bins and assigning weights accordingly.
By applying a weighting loss and clipping the precipitation data to a range of [0, 24], DGMR \citep{ravuri2021} focused on improve performance in forecasting heavy precipitation events.
The tricks help maintain the balance between large and small portions in real-world datasets, contributing to its widespread adoption in predicting precipitation.
Given a predicted future time step $x_t\in \mathbb{R}^{h \times w}$ and the number of elements $N$ in $x_t$, the weighted loss functions are formulated as follows:
\begin{equation}
    \mathcal{L}_{weighted} = \dfrac{1}{N}\sum_{t=m+1}^{m+n}\|(x_t - G(x_t)) \odot w(x_t)\|_{L_1\ or\ L_2},
\end{equation}
where $w(x_t)\in \mathbb{R}^{h \times w}$ denotes the weights corresponding to rainfall and $\odot$ represents matrix multiplication.
Researchers commonly adopted mean squared error (MSE) \citep{shi2017, cambier2023} and mean absolute error (MAE) \citep{ravuri2021, zhang2023} for basis loss function. 
Recent studies used a loss function that combining these metrics improves performance in learning outliers \citep{wu2021, wang2022, ma2024}.

\subsection{Pooling loss}
The purpose of the pooling loss function is to measure the differences between the actual and predicted values in the larger receptive field by max-pooling before applying the objective function. 
Given a number of pixels $N$ of a batch, the learning objective is formulated as follows:
\begin{equation}
    \mathcal{L}_{pooling} = \dfrac{1}{N}\sum_{t=m+1}^{m+n}\|(Q(x_t) - Q(G(x_t)))\|_{1},
\end{equation}
where $Q$ denotes max-pooling. Max-pooling is a type of operation that selects the maximum value within a receptive field from the region of the feature map covered by the filter. The loss function aims to capture how well the model predicts the overall spatial patterns or features in the data rather than focusing on individual data points. This aggregation step helps address the sparse problem and extract the most salient features, which can improve the capture of spatial patterns in various precipitation cases. In other real-world problems, datasets often exhibit long-tail distributions, making the loss function a powerful technique to improve performance on such datasets \citep{rota2017}. Adjusting the max-pooling in the loss function helps the model focus more on the important features.

\subsection{Motion loss}
Precipitation forecasting models frequently exhibit limitations in accurately predicting precipitation spatial scales that refer to the horizontal extent and organization of precipitation systems within a model grid.
For predicting high-resolution precipitation, a motion regularization term was designed to enhance the smoothness of motion fields associated with heavy precipitation events. 
Let $\nabla \text{v}^{1,2} \in \mathbb{R}^{n \times h \times w}$ denote the x- and y-axis gradients obtained by the Sobel filter, and the motion loss is given by:
\begin{equation}
    \mathcal{L}_{motion} = \dfrac{1}{N}\sum_{t=m+1}^{m+n}\underbrace{\|\nabla \text{v}_t^{1} \odot \sqrt{w(x_t)}\|_{2}}_{\text{x-axis velocity $\odot$ density}} + \underbrace{\|\nabla \text{v}_t^{2} \odot \sqrt{w(x_t)}\|_{2}}_{\text{y-axis velocity $\odot$ density}},
\end{equation}
where $N$ is the number of elements. This term incorporates considerations of both the continuity equation and the observed tendency for larger precipitation patterns to persist for longer durations compared to smaller ones.
Larger precipitation patterns like mesoscale convective systems have more moisture and energy available, allowing them to sustain precipitation over a wider area and longer time period \citep{hatsuzuka2021, henny2022}.
For example, about 40\% of days with extreme precipitation over 250 mm in the tropics are associated with long-lived convective systems lasting more than 24 h \citep{roca2020}. This method improves to affect past radar fields to predict the persistence of precipitation events. The regularization term is formulated as the spatial gradient of the motion vector, multiplied by the square root of the precipitation intensity.

\subsection{Nowcasting loss}
The nowcasting loss \citep{ko2022} addresses the class imbalance problem by designing probability distributions from a DL model to approximate each class according to the rainfall threshold. The intrinsic complexities of real-world data often lead to scenarios where certain classes are sparse, creating significant challenges for training models. This issue arises when there is an imbalance in the class distribution within the dataset, potentially biasing the training process towards the majority classes \citep{salcedo2024}. For handling the class imbalance problem, the nowcasting loss function prevents them from being biased toward the majority classes for learning to focus on heavy rain.
Given a threshold $\gamma$ of a class $c$ and a $i$-th element in $x$, $TP(c)$ (true positive), $FP(c)$ (false positive), and $FN(c)$ (false negative) are one-hot matrices indicating whether the value is over $c$ or not. For differentiability, the result transforms the one-hot values into probability distributions as follows:
\begin{equation}
\resizebox{.9\hsize}{!}{$
    \widetilde{TP}_{i}(c),  \widetilde{FN}_{i}(c),  \widetilde{FP}_{i}(c) = \begin{cases}
    P(x_{i,c}), 1-P(x_{i,c}), 0 & \text{if} \ x_{i} \geq  \gamma,\\
    1-P(x_{i,c}), 0, P(x_{i,c}) & \text{otherwise.}
    \end{cases}
$}
\end{equation}
Let the nowcasting loss function be denoted by $\mathcal{L}_{nowcasting}$, the probability between classes can be learned by:
\begin{equation}
\begin{split}
    \mathcal{L}_{nowcasting} = -\dfrac{1}{2} \left( \dfrac{\widetilde{TP}(\text{R})}{\widetilde{TP}(\text{R}) + \widetilde{FP}(\text{R}) + \widetilde{FN}(\text{R})} \right. \\
    \left. + \dfrac{\widetilde{TP}(\text{H})}{\widetilde{TP}(\text{H}) + \widetilde{FP}(\text{H}) + \widetilde{FN}(\text{H})} \right)
\end{split}
\end{equation}
where R and H denote a class defined as the union of light and heavy rain, respectively. 
In this case, for heavy rain, the loss function is calculated twice for the criteria R and H, which mitigates the class imbalance problem as it is trained on a relatively focused basis.

\subsection{Plotting position loss}
The plotting position (PP) loss function \citep{xu2024} is designed to solve the imbalanced regression task for precipitation nowcasting. 
The cost function computes a weight for each sample that is inversely proportional to the probability of occurrence.
This means that samples with rare target values will be given higher weights, while samples with common target values will have lower weights. In traditional loss function, each sample contributes equally to the overall loss function, regardless of how common or rare its target value is. The PP loss is defined by reformulating Eq. \ref{eq:pp1},  \ref{eq:pp2}, and \ref{eq:pp3} as follows:
\begin{equation}\label{eq:pp1}
    \mathcal{L}_{pp} = \dfrac{1}{N}\sum_{t=m+1}^{m+n}\underbrace{f_w (x_i)}_{\text{weight function}} \odot \underbrace{\mathcal{L}(\hat{x}_i, x_i)}_{\text{an objective function}},
\end{equation}
\begin{equation}\label{eq:pp2}
    f_w(x_i) = \dfrac{PP(x_i)}{\frac{1}{N}\sum_{t=m+1}^{m+n} PP(x_i)},
\end{equation}
\begin{equation}\label{eq:pp3}
    PP(x_i) = \dfrac{1 + \left|\left\{\sum_{j=1}^{c} s_j : s_j < s_i\right\}\right|}{N+1},
\end{equation}
where $s_j$ denotes the number of elements in each class of the precipitation dataset, $s_i$ represents the number of elements in $x_i$, and $r_k$ denotes the ranking function. $r_k(x_t)$ denotes the sequence of permutations containing the ordering of the $i$-th element in $x_i$. $N$ denotes the number of elements in $x_i$, and $s_i$ represents the number of elements in the interval where the elements of $x_t$ are contained. $f_w$ represents a weight function for the rarity of the target value. This formula ensures that rare data points have a greater impact on the loss and gradient calculations, helping the model to better learn from these important but scarce samples.

\section{Evaluation metrics}
\label{sec:evaluation}
The evaluation metrics play a critical role in evaluating the performance of the proposed method.
As the number of precipitation forecasting models increases, it is necessary to evaluate different algorithms to verify the models uniformly using common evaluation metrics.
This section introduces benchmark evaluation metrics commonly used in precipitation forecasting.
Table \ref{tab:eval} shows categorized evaluation metrics for the verification from different perspectives.
This comparative analysis helps in understanding which models are most suitable for specific forecasting tasks and provides valuable insights for further research and development efforts.

\subsection{Global-level accuracy} 
The evaluation metrics for the global-level assessment determine how close forecasts are to the observation data for all the pixels, playing an important role in the overall performance verification of the forecasts. The mean square error (MSE) \citep{shi2015} and mean absolute error (MAE) \citep{shi2017} are important metrics for evaluating the overall frames of a model. A small value indicates that the prediction model performed better. The Pearson correlation coefficient (PCC) \citep{ravuri2021} is widely used as a metric to measure the statistical links between observational and predictive results.

\begin{table}[t]
    \centering
    \begin{threeparttable}
    \renewcommand{\arraystretch}{1.2}{
    \begin{tabular}{c|c}
    \Xhline{2\arrayrulewidth}
    \textbf{Category} & \textbf{Metrics} \\
    \hline
    Global-level & MAE, \textbf{MSE}, PCC \\
    \multirow{2}{*}{Binary\tnote{†}} & Accuracy, \textbf{CSI}, FAR, FSS, F1-score,\\
    & POD, Precision, HSS, Recall \\
    Downscale &  CSI-neighborhood, \textbf{FSS}, Pooled CRPS\\
    Sharpness & GDL, LPIPS, PSNR, \textbf{SSIM}\\
    \Xhline{2\arrayrulewidth}
    \end{tabular}}
    \begin{tablenotes}
    \item[†] \small {Rainfall thresholds (mm): 0.2, \textbf{0.5}, 1, 2, 4, 8, 10, 20
    \item[] \small \ \ Reflectivity thresholds (dBZ): 20, \textbf{30}, 35, 40, 50
    \item[] \small \ \ Image thresholds (0--255): 16, 74, 133, 160, 181, 219}
    \end{tablenotes}
    \end{threeparttable}
    \vspace{-1em}
    \caption{Overview of the evaluation metrics for precipitation. \textbf{Bold} font signifies the most frequently used metrics in our survey literature. In binary accuracy, the average point of thresholds is also used as the criterion for verification.}
    \label{tab:eval}
\end{table}

\subsection{Binary accuracy} 
Binary accuracy verifies the pixel-by-pixel precipitation intensity using confusion metrics based on rainfall thresholds. The key to verifying the forecasting model is to evaluate at each pixel where and to what extent precipitation will fall. This verification process involves the use of confusion metrics, which are calculated based on predefined rainfall thresholds. The rainfall thresholds are typically defined in association with the distribution of rainfall intensity. Representative examples include the critical success index (CSI), accuracy, F1-score (F1), precision, recall, probability of detection (POD), equivalent threat score (ETS), and false alarm ratio (FAR) \citep{ravuri2021, trebing2021, ma2024}. The point to note is that, in the case of precipitation, instead of evaluating on a mini-batch basis, one should compute the confusion matrix for the entire batch, taking into account cases where there is no rain (TP=0).

\subsection{Downscale accuracy} 
Both \textit{global accuracy} and \textit{binary accuracy} are assessed on a pixel-level basis, such that even a slight spatial deviation will result in a penalty. Thus models will tend to unrealistically smooth out with increasing lead time in order to minimize the loss while sharp predictions will be over-penalized despite their capability of preserving structural coherency with observations.  For non-blurring models like GANs, alternate evaluation metrics such as the fraction skill score (FSS) \citep{an2023} are more appropriate which evaluates the model by expanding the spatial scale. The CSI-neighborhood \citep{zhang2023} or pooled continuous ranked probability score (CRPS) \citep{ravuri2021}, measured at a lower spatial scale, is used for the same purpose.

\subsection{Sharpness accuracy} 
Sharpness refers to the sensitivity in generating realistic predictions, which is a crucial criterion for evaluating both the boundary quality of precipitation areas and the blurring issue. Given that extreme precipitation events often entail small-scale convective features, the sharpness at which an image is generated serves as an important metric for evaluating the quality of generated images. The degree of blurring of DL models was evaluated using the gradient difference loss (GDL) \citep{wu2021}, learned perceptual image patch similarity (LPIPS) \citep{yu2023}, peak signal-to-noise ratio (PSNR) \citep{wu2023}, and structural similarity index (SSIM) \citep{wu2023, ning2023}. However, they have a limitation on whether sharpness reflects physical patterns, thereby we should complement it with other evaluation techniques to capture the performance.

\section{Recursive strategy}\label{sec:recursive_strategy}
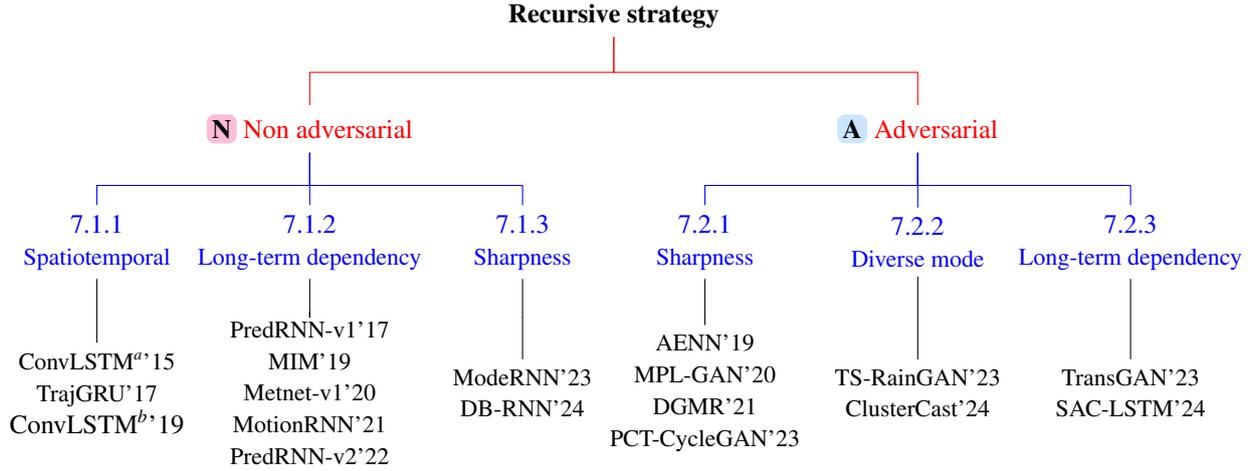
\begin{figure*}[tb]
    \centering
    \begin{tikzpicture}
    [
        level 1/.style = {red, sibling distance = 8cm},
        level 2/.style = {blue, sibling distance = 2.8cm},
        level 3/.style = {black, level distance =2cm},
        edge from parent fork down
    ]
    \node {\textbf{Recursive strategy}}
        child {node {\badge{N} Non adversarial}
        child {node [align=center]{\ref{subsec:dl1} \\ \small Spatiotemporal}
        child {node [align=center]{\small $\text{ConvLSTM}^{a}$’15\\ \small TrajGRU'17\\ $\text{ConvLSTM}^{b}$’19}}}
        child {node [align=center]{\ref{subsec:dl2} \\ \small Long-term dependency}
        child {node [align=center]{\small PredRNN-v1'17\\ \small MIM'19 \\ \small Metnet-v1'20\\ \small MotionRNN'21 \\ \small PredRNN-v2'22}}}
        child {node [align=center]{\ref{subsec:dl3} \\ \small Sharpness}
        child {node [align=center]{\small ModeRNN'23 \\ \small DB-RNN'24}}}}
        child {node {\badge[blue1]{A} Adversarial}
        child {node [align=center]{\ref{subsec:dl4} \\ \small Sharpness}{
        child {node [align=center]{\small AENN'19\\ \small MPL-GAN'20\\ \small DGMR'21 \\ \small PCT-CycleGAN'23}}}}
        child {node [align=center]{\ref{subsec:dl5} \\ \small Diverse mode}
        child {node [align=center]{\small TS-RainGAN'23 \\ \small ClusterCast'24}}}
        child {node [align=center]{\ref{subsec:dl6} \\ \small Long-term dependency}
        child {node [align=center]{\small TransGAN'23 \\ \small SAC-LSTM'24}}}};
    \end{tikzpicture}
    \caption{\textbf{Overview of the recursive strategy.} Methods in the recursive strategy can be categorized into non-adversarial and adversarial. We group the applications based on the above category and then order them chronologically. \badge{N} effectively learn the temporal dependency using recurrent frameworks. \badge[blue1]{A} realistically predict future frames based on GANs. The subcategories were classified based on the core keywords intended to address previous issues in precipitation nowcasting.}
    \label{fig:hist_recursive}
\end{figure*}

In the following sub-section, we provide further details regarding the DL models according to recursive strategy. 
Here, the goal is to examine and provide an overview of recent advances and discoveries in precipitation forecasting.
We include publications that propose DL techniques to address precipitation nowcasting for understanding patterns and trends of state-of-the-art models, even if they have not undergone a peer-review process.
We will chronologically review the technical challenges and their corresponding solutions for each application. The DL models with recursive strategy discussed in the following subsections are organized in Figure \ref{fig:hist_recursive}.

\begin{figure*}[tb]
    \centering
    \begin{subfigure}{\textwidth}
        \centering
        \begin{tikzpicture}
            \node at (0,0) {\includegraphics[width=0.8\linewidth]{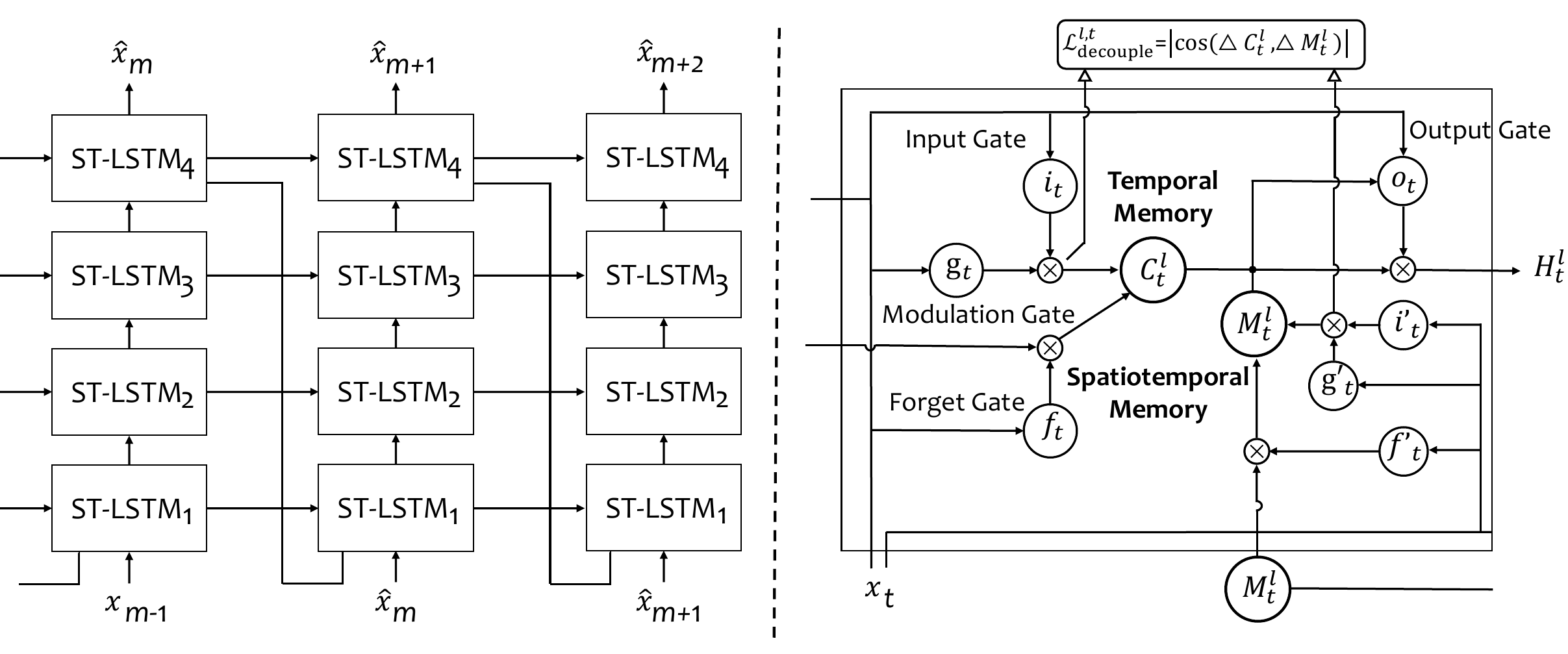}};
        \end{tikzpicture}
        \caption{Schematic of the non-adversarial-based PredRNN2 \citep{wang2022}. (Left) The PredRNN2 overview. (Right) The structure of an ST-LSTM cell. (Source) The figure is adapted from the original paper. $\hat{x}$, $m$, and $t$ represent a predicted frame, input frames, and any time step, respectively. \cite{wang2022} adds the convolution layer upon the increments of $C^l_t$ and $M^l_t$ (memory bank) at every time step, and employs a decoupling loss to explicitly extend the distance between them in latent space to focus on different aspects of spatiotemporal variations.}
        \label{fig:predrnn}
    \end{subfigure}
    \begin{subfigure}{\textwidth}
        \centering
        \begin{tikzpicture}
            \node at (0,0) {\includegraphics[width=0.75\linewidth]{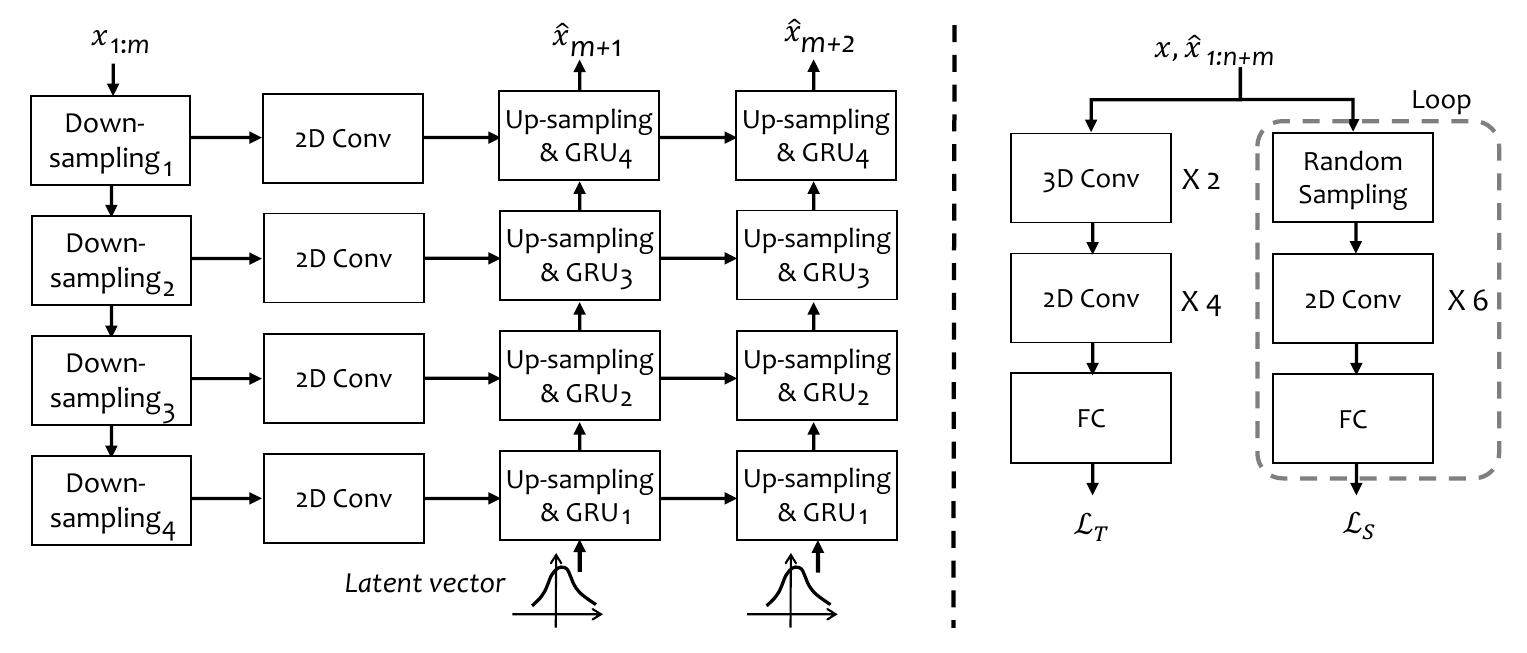}};
        \end{tikzpicture}
        \caption{Schematic of the adversarial-based DGMR \citep{ravuri2021}. (Left) A generator with a latent vector. (Right) Discriminators. The conditions of precipitation observations are represented from the input frames and applied to the latent vector by the generator for predicting future time steps. For the temporal discriminator, 3D convolution is used to learn the temporal distribution. To learn the spatial distribution, a single frame is randomly selected multiple times from the frames, and 2D convolution is applied in each loop.}
        \label{fig:dgmr}
    \end{subfigure}
    \caption{Examples of the model architectures with the recursive strategy}
    \label{fig:framework_recursive}
\end{figure*}

\textbf{Advantages and disadvantages.} (\textbf{+}) efficiently estimates movement by characterizing the stochastic dependency in time series, (\textbf{-}) but is computationally intensive and suffers from accumulative error over longer forecasting time.\\
The recursive strategy is commonly used for predicting hazardous weather over a very short term and has been studied with a focus on predicting spatiotemporal motion since leading to biased and suboptimal forecasts, especially for longer forecast lead time. However, the advantage of the recursive forecasting strategy is that it can effectively capture temporal dependency in the time series, which is commonly composed of recurrent neural networks \citep{shi2015, wang2017}.
The strategy can be classified into two types:
\begin{enumerate}
    \item[\textbf{(1)}] \textbf{Non-adversarial-based methods.} (+) efficiently capture space-time dependency, (-) but tend to become increasingly blurred with longer lead time.
    \item[\textbf{(2)}] \textbf{Adversarial-based methods.} (+) show strength in generating future frames with sharpness, (-) but are comparatively prone to instability, encountering the mode collapse problem.
\end{enumerate}
 Figure \ref{fig:framework_recursive} illustrates two distinct network architectures, each representing a different category within the defined structures.  The figure offers a visual understanding of the architectures employed in the categorized structures to facilitate the comparison between the two different structures.

\subsection{Non-adversarial-based methods}

\subsubsection{Spatiotemporal}\label{subsec:dl1}
First appearing in 2015, $\text{ConvLSTM}^{a}$ \citep{shi2015} was presented as a network for predicting space-time patterns by applying convolutions to the recurrent state transitions of a long-short term memory (LSTM) cell.
With the $\text{ConvLSTM}^{a}$ showing better performance than traditional statistical models in short-term precipitation forecasting within 2 h, DL-based systems have attracted the attention of researchers. As the network computes the weights in cells with fixed resolution, there is a limitation to learning the non-linear representations of motion flow. TrajGRU \citep{shi2017} and $\text{ConvLSTM}^{b}$ \citep{tran2019} have been proposed as hierarchical convolution recurrent neural network (ConvRNN) cells capable of better predicting varying motions and learning nonlinear patterns. This hierarchical learning helps capture intricate moving patterns and represent diverse physical patterns. ConvRNN cells process recursively, one frame at a time, allowing them to incorporate the order and context of the data into their networks. The sequential structure enables the networks to understand and capture the spatiotemporal dependency between frames.

\subsubsection{Long-term dependency}\label{subsec:dl2}
The disadvantage of the recursive strategy lies in long-term dependency. Chaotic dynamics in temporal rainfall transform the joint distributions over time. This results in complex non-stationarity of the precipitation data, manifesting in the variability of rainfall patterns across different temporal scales. To address the intrinsic variation in consecutive contexts,  MetNet-v1 \citep{sonderby2020} and ModeRNN \citep{yao2023} introduced ConvRNN cells with an attention mechanism, a framework designed to enable the memory state to interact between the states. Designing attention modules helps aggregate large contexts and interact with the input state and the previous hidden state. From inside a memory cell, there have been studies on binding cells to capture non-stationarity by memorizing the top layer from each time step \citep{wang2017, wang2019, wu2021}. For example, RNN cells interconnected in a cascaded structure \citep{wang2019} are used to handle space-time non-stationarity and capture differential information, or a motion-highway method \citep{wu2021} is presented, which decomposes transient variations and motion trends. PredRNNs \citep{wang2017, wang2022} learned the non-stationarity of deformations within time steps by adopting nonlinear neurons between time-adjacent RNN states. PredRNN-v2 \citep{wang2022} addressed the issue of accumulated error in long-term forecasting, proposed through reverse scheduled sampling, which learns the long-term dynamics from historical observations by randomly hiding real observations with decreasing probabilities. The PredRNN-v2 structure is shown in Figure \ref{fig:predrnn}, where the spatiotemporal cells can enhance increments of memory states between RNN cells.

\subsubsection{Sharpness}\label{subsec:dl3}
The predicted future frames from the non-adversarial-based models tend to blur as the forecast lead times increase. The MSE function is commonly used for precipitation nowcasting, but such global-level loss functions have some limitations that can lead to blurry outputs. When minimizing MSE loss, the model tends to generate overly blurry predictions by averaging out high-frequency details of precipitation. To address the blurring issue, the DB-RNN \citep{ma2024} involves two steps: forecasting network (MS-RNN) and deblurring network (DB-RNN) with skip connections. The deblurring mechanism can mitigate autoregressive error accumulation, enhancing frame-by-frame deblurring effectiveness over time. In order to optimize the networks, they added regularization terms of GDL and cross-entropy (CE) loss to enhance prediction clarity and partially mitigate blurring. Another approach to mitigate blurring issues is to learn diverse representations across various types of precipitation. While mode collapse might not directly impact blurring in regression tasks, ModeRNN \citep{yao2023} contributes to generating realistic predictions by uncovering the compositional structures of latent modes. The authors introduced adaptive fusion weights, which disentangle features within each cell, facilitating a wide spectrum of mode representations and adaptive weighting to accommodate different spatiotemporal dynamics.

\subsection{Adversarial-based methods}
\subsubsection{Sharpness}\label{subsec:dl4}
The outputs predicted by recursive strategy tend to blur with increasing forecast times \citep{whang2022} when DL models are trained with global-level loss functions, e.g., MAE or MSE. While some approaches may reduce the extent of blurriness, it remains challenging to generate outputs that sharply represent real-world scenarios. Several studies have demonstrated the potential of generative adversarial networks (GANs) \citep{creswell2018} by solving the blurriness problem and demonstrating reliable predictive performance \citep{kupyn2018, brock2018}. Many studies have presented GAN-based methods that realistically predict future frames, including vanilla GAN \citep{jing2019}, Conditional GAN \citep{liu2020}, and CycleGAN \citep{choi2023}. Upon analyzing the commonalities among most GAN models, it is observed that they typically consist of two discriminators to learn the spatiotemporal distribution. The development of adversarial-based models, especially DGMR \citep{ravuri2021} showed promising results by utilizing a hierarchical ConvGRU generator and mapping multivariate Gaussian noise to precipitation fields, as illustrated in Figure \ref{fig:dgmr}. The DGMR tends to show low variance even with perturbations added to the input frames \citep{leinonen2023}. This suggests that the DGMR can effectively learn precipitation physics representations and reliably predict future data, despite learning the spatiotemporal distribution from random noise. 

\subsubsection{Diverse mode}\label{subsec:dl5}
Many GANs suffer from mode collapse due to instability during the adversarial training process, caused by the discriminator becoming too powerful or issues with gradient updates \citep{wiatrak2019}. The mode collapse refers to the condition where a generator fails to capture the diversity mode of the data and produces highly similar, representing only a limited subset of the true data distribution. Precipitation is influenced by the complex interaction of regional, seasonal, and even plant and soil types, and heterogeneous learning and dynamic factors contribute to the temporal variability of precipitation distribution \citep{shi2022}. When predictive models are applied to precipitation datasets with highly mixed dynamics, the DL models often constrain diverse space-time deformations \citep{yao2023}. TS-RainGAN \citep{wang2023} is a task-segmented framework designed to improve heavy rainfall prediction by capturing the complexities of precipitation systems. They proposed two distinct models: one predicts spatiotemporal features to understand rainfall evolution, and the other generates high-quality images for accurate heavy rain detail capture. This approach enhances predictive accuracy and detailed visualization, preventing GAN mode collapse. ClusterCast \citep{an2024} introduced a GAN model with a self-cluster network for forecasting precipitation and capturing transient movement changes. Learning representations of distinct modes improves flexibility across precipitation types, further aiding in stable representation.

\subsubsection{Long-term dependency}\label{subsec:dl6}
Adversarial-based models with the recursive strategy struggle to capture long-term dependency due to the vanishing and exploding gradient problem. SAC-LSTM \citep{she2023} aggregates spatiotemporal features and captures the representation of input over many time steps by applying an attention module in the LSTM cell. After predicting future frames, the connected discriminator estimates whether the frames are generated or real by CE loss. While recursive strategy in theory can model strong temporal dependency, in practice various architectural innovations and training techniques are required to effectively address the underlying long-term dependency. ConvLSTM-TransGAN \citep{yu2023b} is a framework that addresses blurring issues while preserving temporal dependency by merging ConvLSTM and TransGAN. TransGAN \citep{jiang2021}, a GAN model built upon the Transformer architecture, enhances feature resolution progressively and incorporates a multi-scale discriminator capable of capturing both high-level semantics and fine-grained textures.

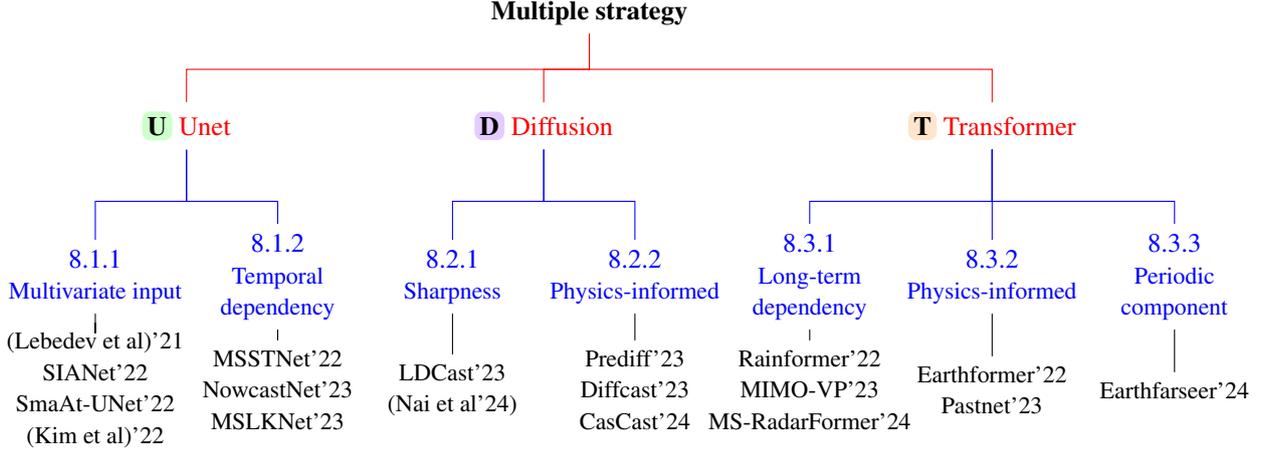
\begin{figure*}[t]
    \centering
    \begin{tikzpicture}
    [
        level 1/.style = {red, sibling distance = 5.3cm},
        level 2/.style = {blue, sibling distance = 2.4cm, level distance =2cm},
        level 3/.style = {black, level distance =1.5cm},
        edge from parent fork down
    ]
    \node {\textbf{Multiple strategy}}
        child {node {\badge[green]{U} Unet}
        child {node [align=center] {\ref{subsec:dl7} \\ \small Multivariate input}
        child {node [align=center] {\small (Lebedev et al)'21\\ \small SIANet'22 \\ \small SmaAt-UNet'22 \\ \small (Kim et al)'22}}}
        child {node [align=center] {\ref{subsec:dl8} \\ \small Temporal \\ \small dependency}
        child {node [align=center]{\small MSSTNet'22 \\ \small NowcastNet'23 \\ \small MSLKNet'23 }}}}
        child {node [xshift=-0.6cm] {\badge[purple1]{D} Diffusion}
        child {node [align=center] {\ref{subsec:dl9} \\ \small Sharpness}
        child {node [align=center] {\small LDCast'23 \\ \small (Nai et al'24)}}}
        child {node [align=center] {\ref{subsec:dl10} \\ \small Physics-informed}
        child {node [align=center]{\small Prediff'23\\ \small Diffcast'23\\ \small CasCast'24}}}}
        child {node {\badge[orange1]{T} Transformer}
        child {node [align=center] {\ref{subsec:dl11} \\ \small Long-term \\ \small dependency}
        child {node [align=center] {\small Rainformer'22 \\ \small MIMO-VP'23 \\ \small MS-RadarFormer'24}}}
        child {node [align=center] {\ref{subsec:dl12} \\ \small Physics-informed}
        child {node [align=center] {\small Earthformer'22 \\ \small Pastnet'23}}}
        child {node [align=center] {\ref{subsec:dl13} \\ \small Periodic \\ \small component}
        child {node [align=center]{\small Earthfarseer'24}}}};
    \end{tikzpicture}
    \caption{\textbf{Overview of multiple strategy.}  Applications are categorized into UNet, Diffusion, and Transformer. Models are characterized based on the methods they report results on and are ordered chronologically. \badge[green]{U} effectively capture channel-wise dependency in multivariate input data. \badge[purple1]{D} realistically predict future frames. \badge[orange1]{T} robust long-term dependency. The subcategories were classified based on the core keywords intended to address previous issues in precipitation nowcasting.}
    \label{fig:hist_multiple}
\end{figure*}

\section{Multiple strategy}\label{sec:multiple_strategy}
This section provides additional details about DL models based on direct multiple strategy. We review publications that reflect trends in state-of-the-art models listed in three frameworks as UNet, Diffusion, and Transformer. We summarize what attempts have been made to solve existing problems in precipitation nowcasting based on each network structure. Figure \ref{fig:hist_multiple} shows a diagram organizing the proposed models for each of the categorized architectures. 

\textbf{Advantages and disadvantages.} (\textbf{+}) robust against such accumulating errors, (\textbf{-}) but leads to difficulties in capturing temporal dependency between each frame.\\
The key advantage of the direct multiple strategy lies in its ability to directly model the entire forecast time, thereby avoiding issues of error accumulation and demonstrating robustness for datasets with high variance over time. This indicates that this approach is better suited for handling the increased uncertainty and complexity of long-term forecasting compared to the recursive strategy. However, this approach may struggle to adequately capture such dependency compared to temporal stochastic prediction models. In this regard, the multiple strategy can provide the disadvantages of learning temporal dependency in short-term trajectory prediction. Multiple strategy can be categorized into three types based on designed architecture:
\begin{enumerate}
    \item[\textbf{(1)}] \textbf{UNet-based methods:} (+) achieve     high-performance for multivariate forecasting, (-) but struggle with learning temporal dependency.
    \item[\textbf{(2)}] \textbf{Transformer-based methods:} (+) address the long-range dependency problem in time series forecasting, (-) but they have computational complexity that scales quadratically with the sequence length.
    \item[\textbf{(3)}] \textbf{Diffusion-based methods:} (+) generate future frames with sharpness and reliability in quantifying uncertainty, (-) but involve sequentially denoising the input over multiple steps is computationally expensive.
\end{enumerate}
Figure \ref{fig:framework_multiple} demonstrates example architectures of three distinct networks within the multiple strategy.

\subsection{UNet-based methods}
\subsubsection{Multivariate input}\label{subsec:dl7}
The UNet structure can efficiently learn the mapping between multivariate input data and radar output frames. Successful forecasting applications based on multivariate input data have been developed \citep{lebedev2019, seo2022}. NowCasting-nets \citep{ehsani2022} compared the performance of UNet-based and RNN-based models for 90 min forecasting using satellite and radar inputs. They found that the UNet-based model outperformed the RNN-based model and that overall performance remained stable over time.
In the Weather4cast competition \citep{gruca2023} for predicting high-resolution radar data from low-resolution satellites and radar data, the winners were models based on 3D-UNet \citep{seo2022,kim2022}. SIANet, the winner of the 2022 competition \citep{seo2022}, focused on radar detection of water droplets larger than 2 mm based on sensor fusing, which are mature cloud particles that may be overlooked during convective initiation. To address this, they proposed a UNet with a large spatial-context aggregation module based on matrix decomposition. Another competitor, presented by \cite{kim2022}, also introduced a UNet with a region-conditioned network that integrates region information into feature maps using orthogonal convolutional layers. In summary, UNet demonstrated strength in learning dependency between variables by effectively fusing information from different sensors.

\subsubsection{Temporal dependency}\label{subsec:dl8}
Deep convolution structure is strong for preserving the global spatial information, but previous UNet models often struggle with non-stationarity and the accompanying variation in data distributions. The UNet model prioritizes analyzing variable correlations over capturing motion trends over time, which makes it less attractive for radar-based univariate forecasting. 
However, NowcastNet \citep{zhang2023} leveraged a sub-network (Evolution Network) to capture temporal dependency, as described in Figure \ref{fig:nowcastnet}. This mitigates the drawbacks of the multiple strategy and has been reported to outperform DGMR. The sub-network of NowcastNet employs a UNet that estimates changes in motion and intensity of radar echoes. These findings suggest that a multi-strategy approach can be a powerful tool for precipitation nowcasting with univariate data, provided a network can be designed to learn temporal dependency. An alternative method is attention mechanisms that capture the temporal dependency in UNets \citep{sonderby2020, trebing2021, tian2023}. For example, MSLKNet \citep{tian2023} encodes features by stacking time-wise attention modules within a UNet latent space, enabling the model to capture both space-time dependency and global patterns. Combining attention modules into UNet networks avoids iterative computation, as these networks do not rely on recursive modules.

\subsection{Diffusion-based methods}

\subsubsection{Sharpness}\label{subsec:dl9}
GANs often struggle to reach a balance point where both the generator and discriminator cannot improve further, resulting in non-convergence \citep{chen2021}. Researchers have actively utilized diffusion \citep{leinonen2023,nai2024}, which is advantageous for its freedom from discriminators and has shown promising results in various real-world applications \citep{ho2020,song2020}. LDCast \citep{leinonen2023} was the first to apply a latent diffusion model to precipitation forecasting. The model was designed using a two-step approach comprising a forecast network to predict future time-step frames and a denoising network for a noise estimation. The denoising network refers to the iterative process of gradually removing noise from an input image, which can learn the data distribution and generate diverse and high-quality samples. LDCast aims to reduce uncertainty by utilizing ensemble members for the average precipitation. However, in previous diffusion-based models, sample variations occur during precipitation forecasting. The variance of samples can lead to the selection of samples that deviate significantly from the ground truth, and ensemble methods may degrade performance \citep{yu2023}. While diffusion models excel at generating high-quality samples, they might lack the fine-grained control required for specific tasks or applications. Generative models may deviate from physical behaviors by generating plausible noise or disregarding domain-specific expertise \citep{gao2023}.

\begin{figure*}[htbp]
    \centering
    \begin{subfigure}{\textwidth}
        \centering
        \begin{tikzpicture}
            \node at (0,0) {\includegraphics[width=0.75\linewidth]{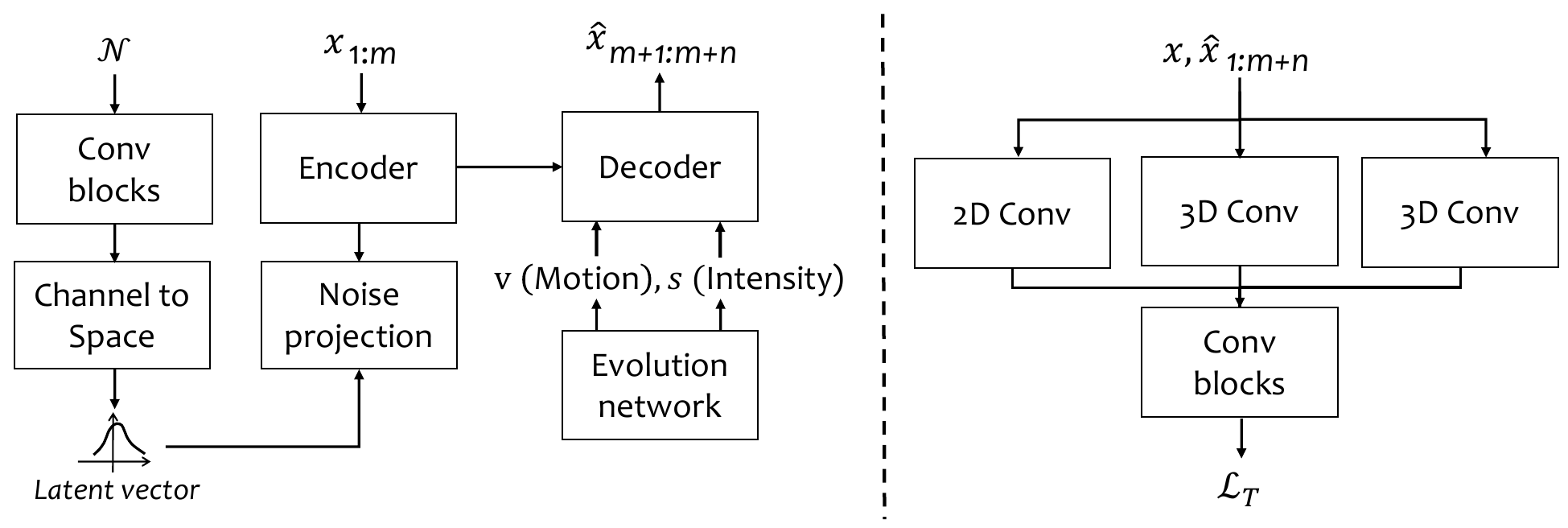}};
        \end{tikzpicture}
        \caption{Schematic of the UNet-based NowcastNet \citep{zhang2023}. (Left) NowcastNet overview based on UNet (Encoder and Decoder) structure. (Right) A temporal discriminator using multiple convolution kernels. The architecture focuses on a UNet-based model designed to learn space-time dependency. This is achieved through an evolution network that enables the model to learn both motion flow and intensity change. By combining these sub-networks, the model can effectively capture the dynamic changes in the data over time and space.}
        \label{fig:nowcastnet}
    \end{subfigure}
    \begin{subfigure}{\textwidth}
        \centering
        \begin{tikzpicture}
            \node at (0,0) {\includegraphics[width=0.8\linewidth]{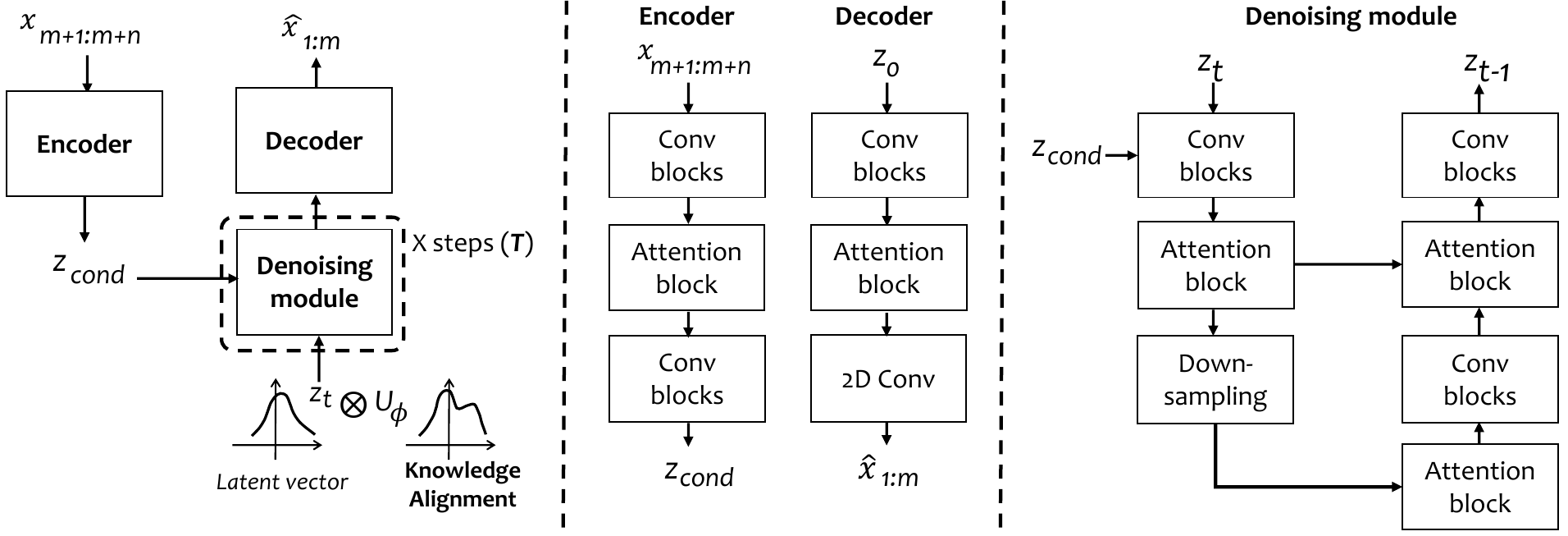}};
        \end{tikzpicture}
        \caption{Schematic of the Diffusion-based Prediff \citep{gao2023}. (Left) Prediff overview during training. (center) Encoder-Decoder structure based on attention mechanism. (Right) Denoising module for estimating observation distribution. Prediff has the advantage of enforcing the spread of the distributions based on knowledge alignment in denoising steps. This model can effectively model the complex, high-dimensional distribution of natural observations without the need for adversarial training, improving the overall learning process and ensuring a more accurate representation.}
        \label{fig:prediff}
    \end{subfigure}
    \begin{subfigure}{\textwidth}
        \centering
        \begin{tikzpicture}
            \node at (0,0) {\includegraphics[width=0.55\linewidth]{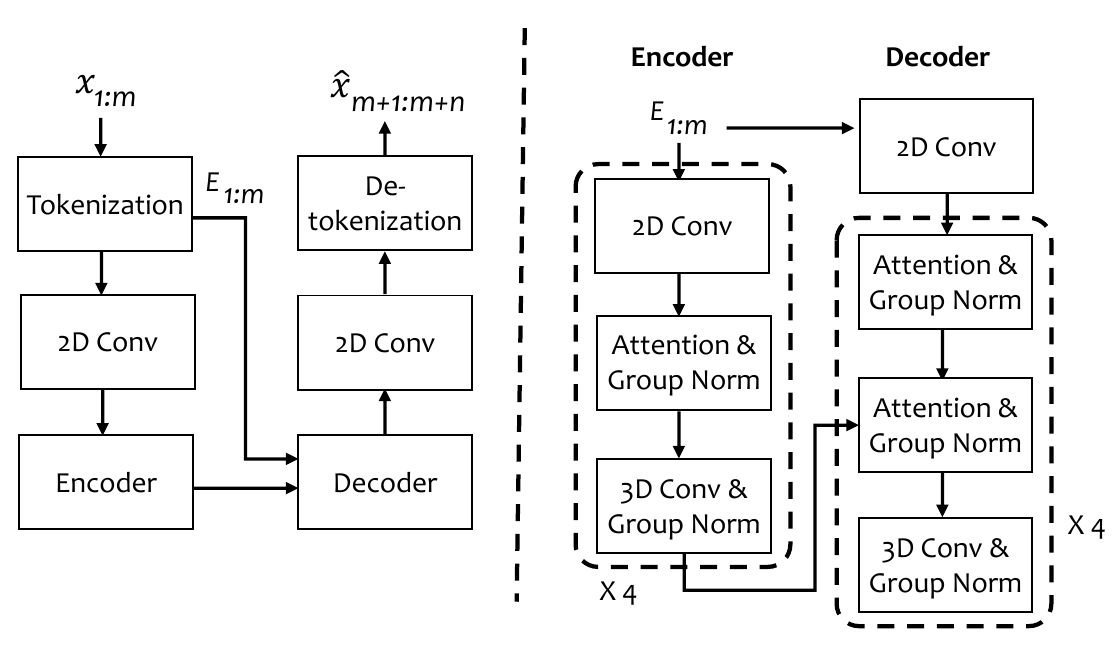}};
        \end{tikzpicture}
        \caption{Schematic of the Transformer-based MIMO-VP \citep{ning2023}. (Left) MIMO-VP overview for predicting future time steps. (Right) Encoder-Decoder structure. The architecture combines convolution with time-wise attention to learn space-time dependency. By leveraging the self-attention mechanism with convolution layers, the model can understand and predict space-time relationships, enabling it to capture complex patterns and dynamics in temporal data. This framework also provides robustness to long-range dependency, allowing the model to effectively capture complex correlations and patterns.}
        \label{fig:mimo}
    \end{subfigure}
    \caption{Examples of the model architectures with the multiple strategy}
    \label{fig:framework_multiple}
\end{figure*}

\subsubsection{Physics-informed}\label{subsec:dl10}
There have been attempts to address existing uncontrolled problems and to predict precipitation with a physics-informed generative model in diffusion-based models. Prediff \citep{gao2023} proposed a network that parameterizes an energy function to adjust transition probabilities, thereby controlling the condition vector during denoising steps. To extract conditional features from the input, Prediff employs the UNet structure of Earthformer \citep{gao2022}. Before denoising steps, the network takes the weather conditions as the basis of noise and aims to minimize discrepancies with the future temporal distribution. Then, the Prediff parameterizes an energy function to adjust the transition probabilities during each denoising step, resulting in predicted future frames that are rich in detail. A simple Prediff framework for enforcing the distribution of images with physical alignment is visualized in Figure \ref{fig:prediff}. DiffCast \citep{yu2023} and CastCast \citep{gong2024} combine deterministic and stochastic diffusion components to capture global and local motion trends. The deterministic component predicts the global pattern, whereas the stochastic diffusion component provides more detail. Configuring only the stochastic model resulted in instability and constrained the capability to decompose precipitation into both global trends and local stochastic components \citep{yu2023}. This indicates that uncontrolled evolution models may struggle to capture the physical distribution of precipitation. Diffcast leverages residual blocks as a diffusion component to effectively exploit multi-scale temporal features. In the case of Cascast, the network emphasizes reducing the complexity of denoising conditioned by a sequence of blurry predictions by adopting an attention mechanism. While diffusion models can be computationally expensive, they can effectively capture intricate patterns present in time series data by estimating perturbations with physics alignment. Diffusion models can enhance representations to preserve coherence when generating long-term forecasts with a high frequency of precipitation and facilitate the modeling of complex temporal distributions.

\subsection{Transformer-based methods}
\subsubsection{Long-term dependency}\label{subsec:dl11}
Transformer-based nowcasting models have achieved remarkable success, demonstrating their ability to effectively capture long-range spatiotemporal dependency in precipitation nowcasting. Transformer efficiently robust long-range dependency by tokenizing data into small patches and learning weights for each patch using self-attention. The self-attention mechanism in the Transformer, while powerful for capturing long-range dependency, may struggle to effectively model short-term local patterns and variations in the data \citep{lin2022}. Rainformer \citep{bai2022} and MS-RadarFormer \citep{geng2024} applied a Swin Transformer by tokenizing time series frames to learn both global and local features. They introduced a fully convolutional network with an attention module that effectively extracts local features from rainfall information, which aids in low-to-medium rainfall intensity prediction. However, precipitation data is to fluid masses spreading in various directions, unlike standard videos which often depict moving objects against a stationary background \citep{fovell1998}. To overcome this drawback, MIMO-VP \citep{ning2023} introduced a spatiotemporal block aimed at learning high-order relationships between time step queries and their corresponding output sequences. As described in Figure \ref{fig:mimo}, the approach involves obtaining sequence-level feature maps using a placeholder embedding method to capture global dependency within the frames. The proposed solution effectively addresses the challenge of global space-time dependency and error accumulation by adopting temporal attention modules with convolution layers. To tackle the permutation-invariant issue inherent to transformers, the network applies temporal positional encoding to input sequences, ensuring smooth and effective processing.

\subsubsection{Physics-informed}\label{subsec:dl12}
Recent studies have strived to enhance model performance by reflecting weather-related knowledge into Transformer-based architectures. Nowformer \citep{park2022} proposed global dynamic attention to learn both local and global information of each frame. For learning the intensity of the moving fluid changes continuously, they proposed Transformer-based global dynamic attention to learn both local and global information of each frame. The Nowformer locally extracts temporal dependency through the local dynamic attention module, facilitating the learning of how the intensity of fluids changes at each point in the frames. Meanwhile, one of the characteristics of weather data is that combining low and high-resolution data can enhance weather forecasting accuracy. By using a combination of low and high-resolution data, weather models can leverage the strengths of each.  The broad context from low-resolution data and the local detail from high-resolution data. Earthformers \citep{gao2022} employed a hierarchical mechanism for multi-resolution attention using image decomposition and connected local and global. They explored the sharing of space-time information in the transformer, as well as designed an efficient computational complexity. MS-RadarFormer \citep{geng2024} proposed a multi-scale patch embedding layer, enabling it to obtain the multiple spatial and temporal scales of the high-resolution precipitation dataset. They applied padding operations to adjust the output dimensions and concatenate embedded tokens. By embedding multi-resolution information, they addressed the issue of underestimating while also achieving slightly sharper predictions for radar output frames.

\subsubsection{Periodic component}\label{subsec:dl13}
As scientific data substantiates the importance of Fourier analysis, it is a well-known fact that the decomposition of complex weather signals into their constituent frequency components can better identify and represent important meteorological phenomena \citep{amon1993}. The Fourier series representation within the spectral elements enables the efficient handling of periodic or quasi-periodic phenomena, such as those encountered in transitional and turbulent flows \citep{fournier2005}. The Fourier-based precipitation nowcasting, such as fast Fourier transform (FFT), can help extract relevant features and improve the performance of forecasting models \citep{wu2023, wu2024}.
The Fourier representation allows the model to capture complex periodic behaviors through the combination of sine and cosine terms. Passing input points through a Fourier feature mapping enables networks to learn high-frequency functions in a low-dimensional setting. The FFT module is utilized to capture high-dimensional nonlinear physical features, enabling the capture of intricate dynamics within systems without relying on differential equation-based nonlinear features \citep{wu2023}. Unlike discrete static frames, the FFT approach transforms data from the continuous time domain to the frequency domain, thereby better preserving long-term dependency in spatiotemporal data \citep{wu2024}. Fourier transforms have gained prominence in the field of precipitation nowcasting, providing a method to discern intricate physical patterns and capture system dynamics without the need for differential equation-based nonlinear features.

\begin{sidewaystable*}
    \centering
    \begin{NiceTabular}{p{4.5cm}C{3em}C{5em}C{5em}C{3em}C{4em}C{2em}*{2}{C{2em}}C{2em}C{8em}}
    \CodeBefore 
    \rowcolor{gray1!25}{1} 
    \rowcolors{3}{gray1!10}{}
    \Body
    \toprule
    \small\textbf{Applications} & \small\textbf{Norm.} & \small\textbf{Loss} & \small\textbf{Data} & \small\textbf{Size} & \small\textbf{Param.} & \small\textbf{\textit{m}} & \small\textbf{\textit{n}} & \small\textbf{\textit{dt}} & \scriptsize\textbf{Z-R} & \scriptsize\textbf{Metrics}\\ 
    \midrule
    \multicolumn{9}{l}{\small{(A) Recursive strategy}}\\
    \midrule
    \scriptsize{\badge{N} $\text{ConvLSTM}^{a}$} \citep{shi2015} \href{https://github.com/ndrplz/ConvLSTM_pytorch}{[link]} & \scriptsize{minmax} & \scriptsize{MSE} & \scriptsize{HKO-7} & \scriptsize{100} & \scriptsize{487 K} & \scriptsize{5} & \scriptsize{15} & \scriptsize{6} & \scriptsize{O} &\scriptsize{MSE, PCC, CSI, FAR, POD}\\ 
    \scriptsize{\badge{N} TrajGRU \citep{shi2017} \href{https://github.com/Hzzone/Precipitation-Nowcasting}{[link]}} & \scriptsize{minmax} & \scriptsize{$\mathcal{L}_{w}$} & \scriptsize{HKO-7} & \scriptsize{480} & \scriptsize{1.9 M} & \scriptsize{5} & \scriptsize{20} & \scriptsize{6} & \scriptsize{O} & \scriptsize{MSE, CSI, HSS}\\
    \scriptsize{\badge{N} PredRNN \citep{wang2017} \href{https://github.com/thuml/predrnn-pytorch}{[link]}} & \scriptsize{minmax} & \scriptsize{MAE, MSE} & \scriptsize{Guangzhou} & \scriptsize{100} & \scriptsize{23.6 M} & \scriptsize{10} & \scriptsize{10} & \scriptsize{6} & \scriptsize{O} & \scriptsize{MSE}\\ 
    \scriptsize{\badge{N} MIM \citep{wang2019} \href{https://github.com/Yunbo426/MIM}{[link]}} & \scriptsize{minmax} & \scriptsize{MSE} & \scriptsize{Guangzhou} & \scriptsize{64} & \scriptsize{28.5 M} & \scriptsize{10} & \scriptsize{10} & \scriptsize{6} & \scriptsize{X} & \scriptsize{MSE, CSI} \\ 
    \scriptsize{\badge{N} Metnet-v1 \citep{sonderby2020} \href{https://github.com/openclimatefix/metnet}{[link]}} & \scriptsize{log} & \scriptsize{-} & \scriptsize{MRMS} & \scriptsize{1024} & \scriptsize{225 M} & \scriptsize{6} & \scriptsize{32} & \scriptsize{15} & \scriptsize{O} & \scriptsize{F1} \\
    \scriptsize{\badge{N} MotionRNN \citep{wu2021} \href{https://github.com/thuml/MotionRNN}{[link]}} & \scriptsize{-} & \scriptsize{MAE, MSE} & \scriptsize{Shanghai} & \scriptsize{64} & \scriptsize{5.2 M} & \scriptsize{5} & \scriptsize{10} & \scriptsize{12} & \scriptsize{X} & \scriptsize{MSE, CSI, GDL, SSIM}\\
    \scriptsize{\badge{N} PredRNN2 \citep{wang2022} \href{https://github.com/thuml/predrnn-pytorch}{[link]}} &  \scriptsize{minmax} & \scriptsize{MAE, MSE} & \scriptsize{Guangzhou} & \scriptsize{128} & \scriptsize{23.8 M} & \scriptsize{10} & \scriptsize{10} & \scriptsize{6} & \scriptsize{X} & \scriptsize{MSE, CSI}\\ 
    \scriptsize{\badge[blue1]{A} DGMR \citep{ravuri2021} \href{https://github.com/openclimatefix/skillful_nowcasting}{[link]}} & \scriptsize{clip} & \scriptsize{$\mathcal{L}_{w}$, Hinge} & \scriptsize{UK, US} & \scriptsize{256} & \scriptsize{98.5 M}  & \scriptsize{4} & \scriptsize{18} & \scriptsize{5} & \scriptsize{O} & \scriptsize{CSI, PCC, PSD, CRPS}\\
    \scriptsize{\badge[blue1]{A} SAC-LSTM \citep{she2023}  \href{https://github.com/LeiShe1/SAC-LSTM-MindSpore}{[link]}} & \scriptsize{minmax} & \scriptsize{MSE, CE} & \scriptsize{CIKM} & \scriptsize{128} & \scriptsize{31.1 M}  & \scriptsize{5} & \scriptsize{10} & \scriptsize{6} & \scriptsize{X} & \scriptsize{CSI, HSS, MSE}\\
    \midrule
    \multicolumn{9}{l}{\small{(B) Multiple strategy}}\\
    \midrule
    \scriptsize{\badge[green]{U} Broad-UNet \citep{fernandez2021} \href{https://github.com/jesusgf96/Broad-UNet}{[link]}} & \scriptsize{binary} & \scriptsize{MSE} & \scriptsize{Netherlands} & \scriptsize{288} & \scriptsize{11 M} & \scriptsize{12} & \scriptsize{6} & \scriptsize{5} & \scriptsize{O} & \scriptsize{MSE, Accuracy, Precision, Recall} \\
    \scriptsize{\badge[green]{U} SmaAt-UNet \citep{trebing2021} \href{https://github.com/HansBambel/SmaAt-UNet}{[link]}} & \scriptsize{minmax} & \scriptsize{MSE} & \scriptsize{Netherlands} & \scriptsize{256} & \scriptsize{4 M} & \scriptsize{4} & \scriptsize{6} & \scriptsize{15} & \scriptsize{O} & \scriptsize{MSE, ACC, CSI, FAR, F1, HSS, Precision, Recall}\\
    \scriptsize{\badge[green]{U} NowcastNet \citep{zhang2023} \href{https://codeocean.com/capsule/3935105}{[link]}} & \scriptsize{clip} & \scriptsize{CE, $\mathcal{L}_{pool}$, $\mathcal{L}_{motion}$}& \scriptsize{China, UK} & \scriptsize{256} & \scriptsize{29.1 M} & \scriptsize{4} & \scriptsize{18} & \scriptsize{10} & \scriptsize{O} & \scriptsize{CSI, PSD}\\
    \scriptsize{\badge[purple1]{D} LDCast \citep{leinonen2023} \href{https://github.com/MeteoSwiss/ldcast}{[link]}} & \scriptsize{log} & \scriptsize{MSE, KL} & \scriptsize{MeteoSwiss} & \scriptsize{256} & \scriptsize{671 M} & \scriptsize{4} & \scriptsize{18} & \scriptsize{5} & \scriptsize{X} & \scriptsize{FSS, CRPS}\\
    \scriptsize{\badge[purple1]{D} Prediff \citep{gao2023} \href{https://github.com/gaozhihan/PreDiff}{[link]}} & \scriptsize{} & \scriptsize{MSE, adv, KL} & \scriptsize{SEVIR} & \scriptsize{128} & \scriptsize{120 M} & \scriptsize{6} & \scriptsize{7} & \scriptsize{10} & \scriptsize{O} & \scriptsize{CSI, Bias}\\
    \scriptsize{\badge[purple1]{D} Diffcast \citep{yu2023} \href{https://github.com/DeminYu98/DiffCast}{[link]}} & \scriptsize{clip} & \scriptsize{MSE, KL} & \scriptsize{SEVIR, MetoNet, Shanghai, CIKM} & \scriptsize{128} & \scriptsize{66.4 M} & \scriptsize{5} & \scriptsize{20} & \scriptsize{5} & \scriptsize{X} & \scriptsize{CSI, HSS, LPIPS, SSIM}\\
    \scriptsize{\badge[orange1]{T} Earthformer \citep{gao2022} \href{https://github.com/amazon-science/earth-forecasting-transformer}{[link]}} & \scriptsize{minmax} &  \scriptsize{MSE} & \scriptsize{SEVIR} &  \scriptsize{384} & \scriptsize{7.6 M} &  \scriptsize{12} & \scriptsize{13} & \scriptsize{5} & \scriptsize{X} & \scriptsize{MSE, CSI}\\
    \scriptsize{\badge[orange1]{T} Pastnet \citep{wu2023} \href{https://github.com/easylearningscores/pastnet}{[link]}} & \scriptsize{-} & \scriptsize{MSE} & \scriptsize{SEVIR} & \scriptsize{384} & \scriptsize{54 M} & \scriptsize{10} & \scriptsize{10} & \scriptsize{5} & \scriptsize{X} & \scriptsize{MAE, MSE, PSNR, SSIM} \\
    \scriptsize{\badge[orange1]{T} MIMO-VP \citep{ning2023} \href{https://github.com/ningshuliang/MIMO-VP}{[link]}} & \scriptsize{-}  & \scriptsize{MAE, MSE} & \scriptsize{China} & \scriptsize{128} & \scriptsize{20.2 M} & \scriptsize{10} & \scriptsize{10} & \scriptsize{-} & \scriptsize{X} & \scriptsize{MAE, CSI, SSIM} \\
    \bottomrule
    \end{NiceTabular}    
    \caption{Overview of applications. We grouped their applications based on their strategy and model architectures. \badge{N}: Non-adversarial, \badge[blue1]{A}: Adversarial, \badge[green]{U}: UNet, \badge[purple1]{D}: Diffusion, \badge[orange1]{T}: Transformer. \textbf{Norm}: data normalization (ref. Table \ref{tab:radar}), \textbf{Loss}: loss function for object functions (adv and KL represent an adversarial loss and Kullback-Leibler divergence, respectively.), \textbf{Data}: Used dataset, \textit{param}: number of parameters, \textbf{Size}: image height \& width, $\bm{m}$: the number of input frames, $\bm{n}$: the number of output frames, $\bm{dt}$: forecasting time interval ($\Delta t$), \textbf{Z-R}: application of the Z-R relationship, \textbf{Metrics}: evaluation metrics. PSD represents radially-averaged power spectral density for analyzing variations in spatial frequency content in an image. The URLs in name tags are publicly accessible codes. Param is written under our reproduction process for cases that are not addressed in their paper. Information not included in the paper is represented by dashes. \textcolor{blue}{[link]} directs to code websites.}
    \label{tab:appl}
\end{sidewaystable*}

\section{Methodology comparison and evaluation}\label{sec:comarison_evaluation}
Leveraging our previous discussion of attributes within an interrelated system of DL models, we compare the properties of DL models in this section. For aiming to review key applications on two strategies, we present two levels of comparison: the methodology-level comparison (Section \ref{sec:comparison_method}) and the performance-level comparison (Section \ref{sec:comparison_performance}).

\subsection{Methodology comparison}
\label{sec:comparison_method}
In this subsection, we survey various applications by drawing upon the insights and methodologies outlined in the preceding sections, thereby providing a structured examination of their respective uses and implications. These categories refer to either the dataset used in the applications, objective function, evaluation metric, or additional information. Table \ref{tab:appl} summarizes the key highlights of these applications. The papers examined in this work were collected from various popular benchmark databases, and state-of-the-art were compared with popular applications with publicly available code.
From the survey, we can derive the following insights: 
\begin{enumerate}
    \item[\textbf{(1)}] \textbf{Preprocessing:} By normalizing the data with min-max normalization and clipping, previous studies aim to reduce the impact of non-normality. Both radar reflectivity and rainfall intensity have been used effectively in precipitation nowcasting, with the choice depending on the specific application and modeling approach. Predicting reflectivity directly can be advantageous as it is the original radar signal, but it requires an additional step to convert from reflectivity using the Z-R relationship for detailed assessment \citep{han2023}.
    \item[\textbf{(2)}] \textbf{Objective function:} One of the most widely used objective functions for precipitation nowcasting is MSE. Recent research indicates that both MAE and MSE are adopted as loss functions. MSE penalizes larger errors more heavily, resulting in smoother nowcasts, while MAE treats all errors equally and can better represent intense precipitation at the cost of higher overall errors \citep{ashesh2022}.
    \item[\textbf{(3)}] \textbf{Evaluation metric:} The MSE is one of the most commonly used metrics for evaluating the performance of precipitation nowcasting models. Several other metrics are also frequently employed to assess different aspects of the nowcasts. SSIM and PSNR are used as evaluation metrics to evaluate the sharpness of predicting frames. CRPS, which evaluates the entire predictive distribution, is also used in nowcasting models. The choice of metrics depends on the specific application requirements and characteristics of the precipitation events being forecasted.
\end{enumerate}
While significant progress has been achieved in precipitation forecasting, the absence of an established baseline makes a limitation of the ability to conduct sophisticated model comparisons. We hope that our study contributes to the dissemination of technology and the revitalization of research by providing a comparison of various applications.

\begin{table*}[tb]
\scriptsize
\centering
\begin{tabular}{p{3.4cm}C{6em}C{5em}*{2}{C{3.2em}}C{4em}C{6em}C{8em}}
    \toprule
    \textbf{Applications} & \multicolumn{4}{c}{\textbf{Moving MNIST}} & \multicolumn{3}{c}{\textbf{Weather data}} \\ 
    \cmidrule(l){2-5} \cmidrule(l){6-8}
    & \textbf{\#Param. (M)} & \textbf{FLOPS (G)} & \textbf{MSE $\downarrow$} &  \textbf{SSIM $\uparrow$} & \textbf{Dataset} &  \textbf{MSE ($10^{-3}$) $\downarrow$} & \textbf{CSI (dBZ) $\uparrow$} \\
    \midrule
    Persistence & & & & & SEVIR & 11.53 & 0.261 (M)\\
    \cdashline{1-8}
    \badge{N} $\text{ConvLSTM}^{a}$ \citep{shi2015} & 14.0 & 30.1 & 182.9 & 0.707 & SEVIR & 9.76 & 0.354 (30) \\
    \badge{N} TrajGRU \citep{shi2017} & - & - & 103.3 & 0.713 & SEVIR & 8.92 & 0.357 (30)\\
    \badge{N} PredRNN \citep{wang2017} & 23.85 & 115.9 & 56.8 & 0.867 & SEVIR & - & 0.404 (M)\\
    \badge{N} MIM \citep{wang2019} & 37.37 & 181.7 & 101.1 & 0.910 & Guangzhou  & 27.8 & 0.429 (30) \\
    \badge[green]{U} UNet \citep{veillette2020} & 16.6 & 0.9 & 110.4 & 0.617 & SEVIR  & - & 0.359 (M) \\ 
    \badge{N} MotionRNN \citep{wu2021} & 7.01 & 49.5 & 25.1 & 0.920 & Guangzhou & - & 0.678 (30)\\
    \badge[blue1]{A} DGMR \citep{ravuri2021} & - & - & - & - & SEVIR & - & 0.268 (30) \\
    \badge[green]{U} MSSTNet \citep{ye2022} & - & - & 21.4 & 0.953 & - & - & -\\
    \badge{N} PredRNN2 \citep{wang2022} & 23.86 & 116.6 & 48.4 & 0.891 & SEVIR & 3.9 & 0.480 (M)\\
    \badge{N} MS-RNN \citep{ma2022} & 5.6 & 163.7 & 46.6 & 0.892 & Germany & - & - \\
    \badge[orange1]{T} Rainformer \citep{bai2022} & 19.2 & 1.2 & 85.83 & 0.730 & SEVIR & 4.0 & 0.366 (M) \\
    \badge[orange1]{T} Earthformer \citep{gao2022} & 7.6 & 34 & 41.79 & 0.896 & SEVIR & 3.7 & 0.442 (M)\\
    \badge{N} ModeRNN \citep{yao2023} & 3.15 & - & 44.7 & 0.898 & Guangzhou & 65.1 & 0.428 (30)\\
    \badge[purple1]{D} Prediff \citep{gao2023} & - & - & - & - & SEVIR  & - & 0.407 (M) \\
    \badge[purple1]{D} Diffcast \citep{yu2023} & - & - & - & - & SEVIR  & - & 0.282 (10, 21, 33)\\
    \badge[orange1]{T} Pastnet \citep{wu2023} & - & - & 31.77 & - & SEVIR & 6.6 & - \\
    \badge[orange1]{T} MIMO-VP \citep{ning2023} & 20.2 & - & 17.7 & 0.964 & China & - & 0.400 (30)\\
    \badge[orange1]{T} Eartgfarseer \citep{wu2024} & - & - & 14.9 & - & SEVIR & 2.8 & 0.471 (M)\\
    \badge[purple1]{D} CasCast \citep{gong2024} & - & - & - & - & SEVIR & - & 0.440 (M)\\
    \bottomrule
\end{tabular}  
\caption{Comparison of experimental results on benchmark data. The weather data is organized into 6 min intervals for predicting 10 frames (1 h). FLOPS, or floating point operations per second, are measured to quantify and compare the computational cost of different models and their resource usage. The SSIM function refers to the structural similarity index measure for assessing the human visual perception and similarity between two images. (M) represents the average score of threshold values 16, 74, 133, 160, 181, and 219. \badge{N}: Non-adversarial, \badge[blue1]{A}: Adversarial, \badge[green]{U}: UNet, \badge[purple1]{D}: Diffusion, \badge[orange1]{T}: Transformer.}
\label{tab:result1}
\end{table*}

\subsection{Performance evaluation}\label{sec:comparison_performance}
To evaluate the performance of models across different settings, we defined two benchmark datasets in this study: Moving the MNIST dataset \citep{srivastava2015} and SEVIR dataset \citep{veillette2020}. The reason for selecting the two benchmark datasets is that they are widely used and among the most commonly used in precipitation nowcasting. The SEVIR dataset offers the advantage of enabling comparisons of computational complexity. Moving MNIST is a synthetic dataset with grayscale images for video prediction. The benchmark dataset consists of 10 frames input and 10 frames output to compare the nowcasting models. We summarized the performance of applications that include at least one of these datasets chronologically in Table \ref{tab:result1}. Note that we included cases where the code was officially released and experiments comparing performance on the same data were conducted in other studies. For the threshold of CSI, comparisons were written based on the average, and in cases where the average value was not provided, we used the most commonly used threshold of 30 as the reference. We set up the criteria for comparing computational complexity and model performance using the Moving MNIST dataset, based on the number of parameters, FLOPS, MSE, and SSIM.

\begin{figure*}[tb]
\scriptsize
\centering
\begin{tikzpicture}
    \node at (-0.6,0) [above=8.2em] {\scriptsize{\badge{N} MIM \citep{wang2019}}};
    \node at (5.5,0) [above=8.2em] {\scriptsize{\badge[green]{U} RainNet \citep{ayzel2020}}};
    \node at (11.6,0) [above=8.2em] {\scriptsize{\badge[blue1]{A} DGMR \citep{ravuri2021}}};
    \node at (-0.6,-2.9) {\scriptsize{\badge{N} Metnet-v2 \citep{espeholt2022}}};
    \node at (5.5,-2.9) {\scriptsize{\badge[orange1]{T} Rainformer \citep{bai2022}}};
    \node at (11.6,-2.9) {\scriptsize{\badge[purple1]{D} \citep{nai2024}}};
    \node (image1) at (-0.6,0) {\includegraphics[width=0.31\textwidth]{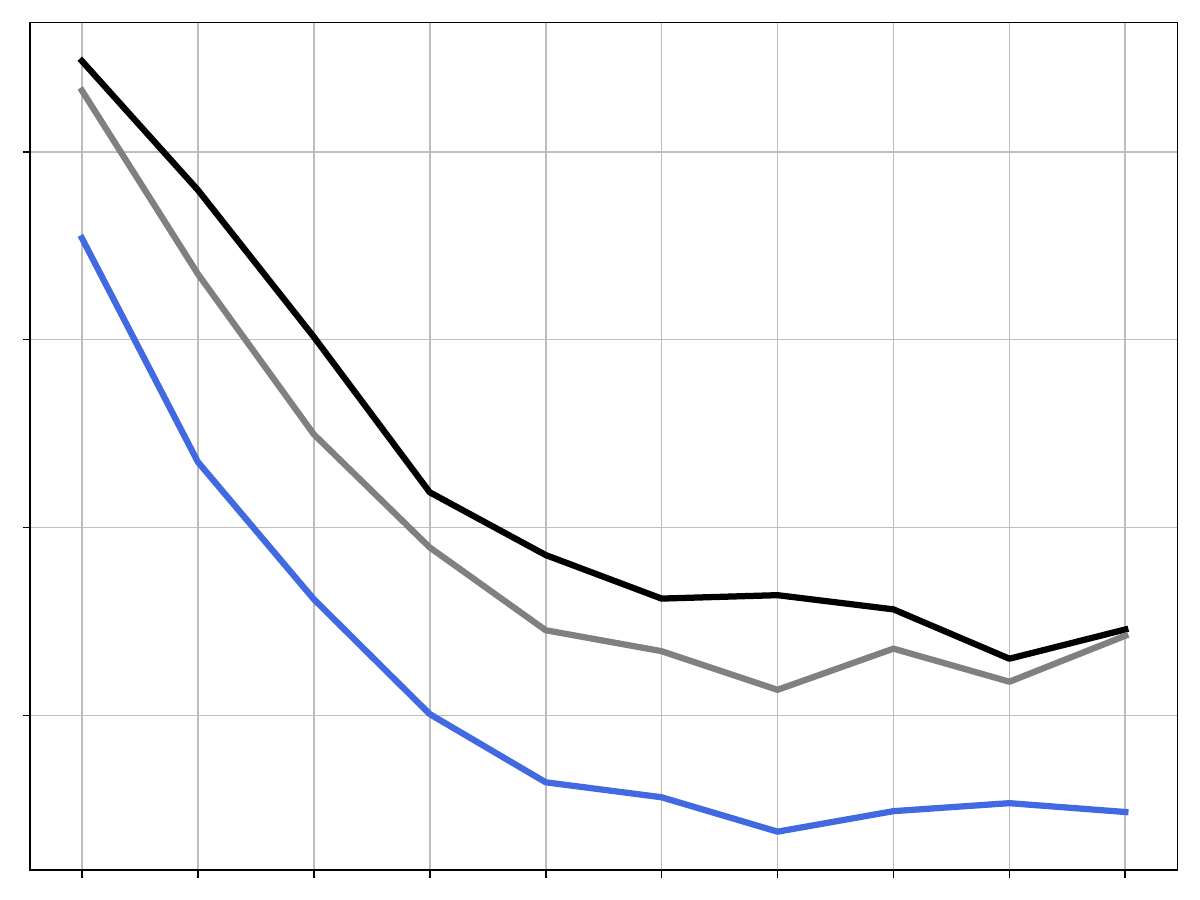}};
    \node (image2) at (5.5,0) {\includegraphics[width=0.31\textwidth]{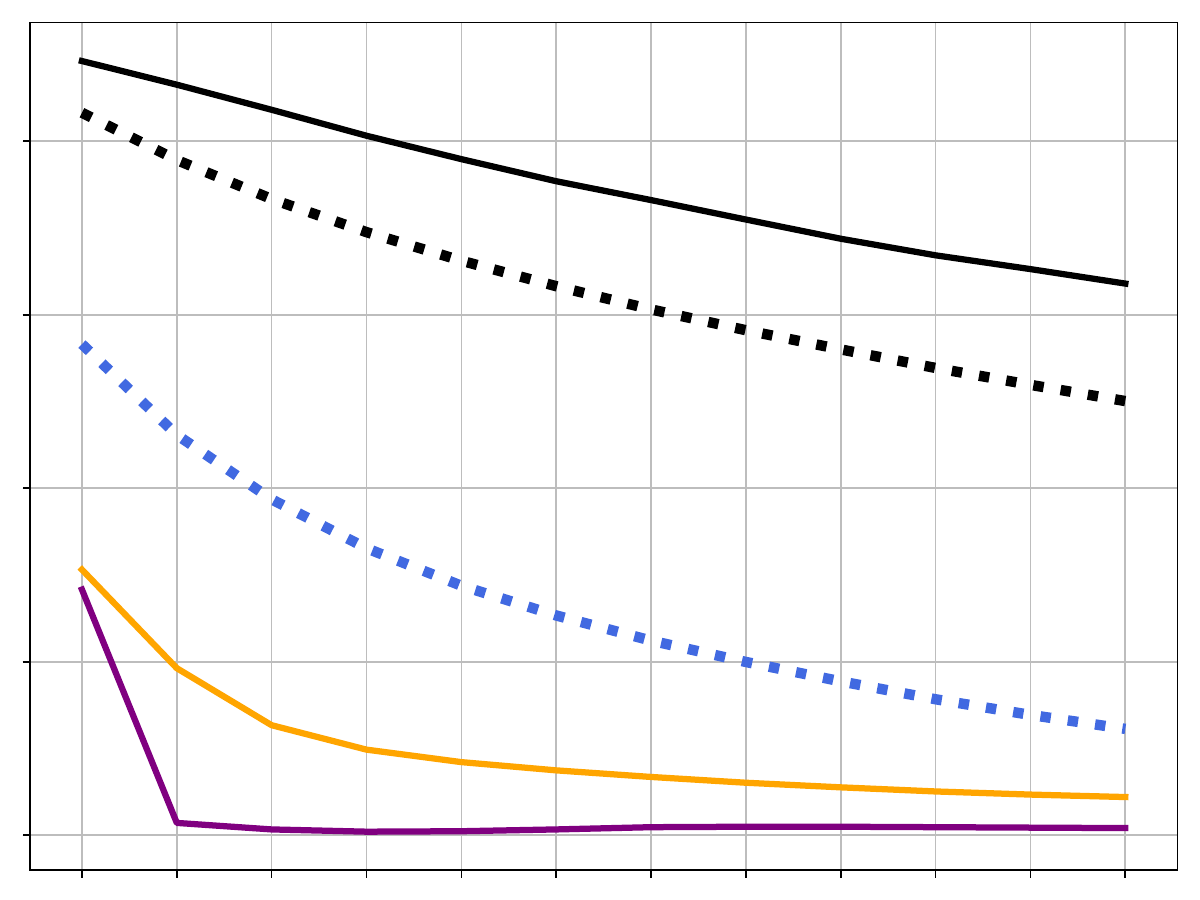}};
    \node (image2) at (11.6,0) {\includegraphics[width=0.31\textwidth]{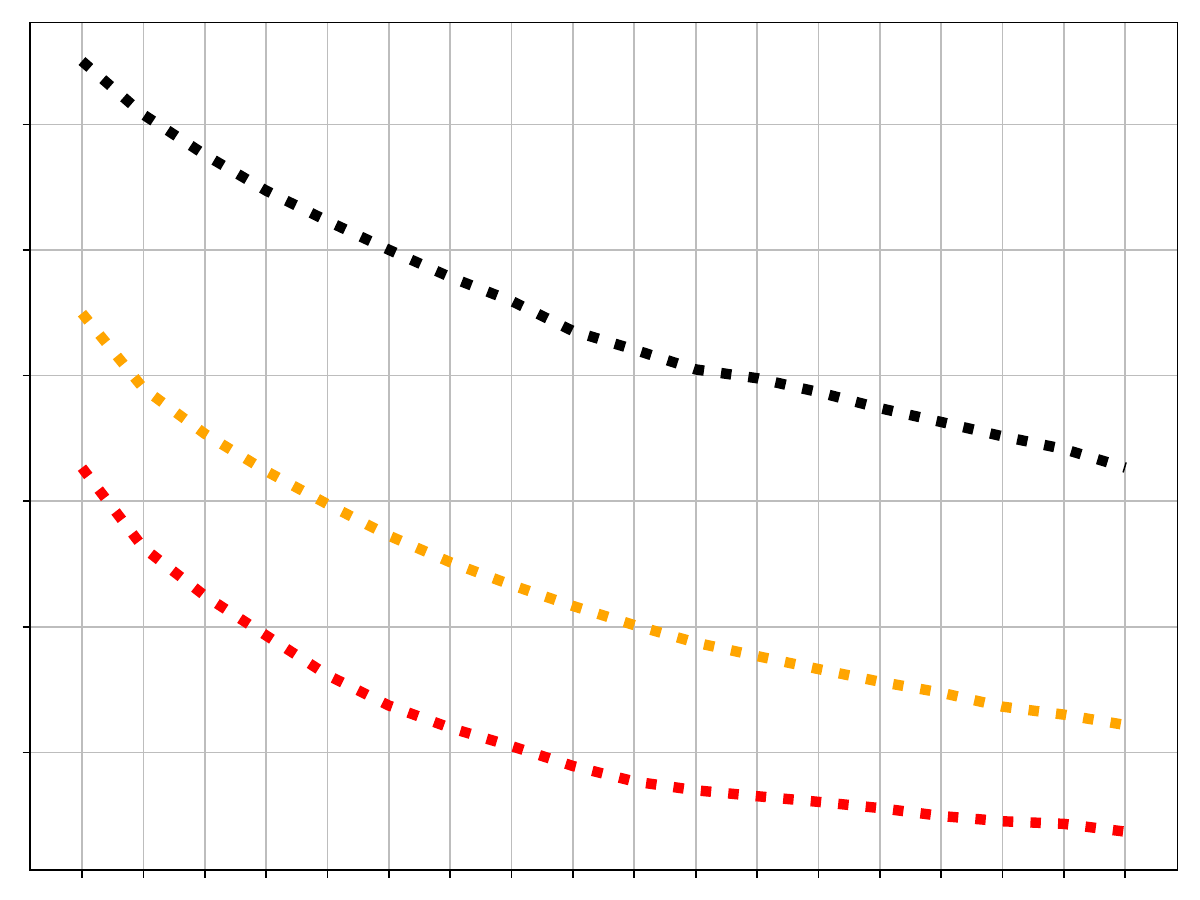}};
    \node (image1) at (-0.6,-5.2) {\includegraphics[width=0.31\textwidth]{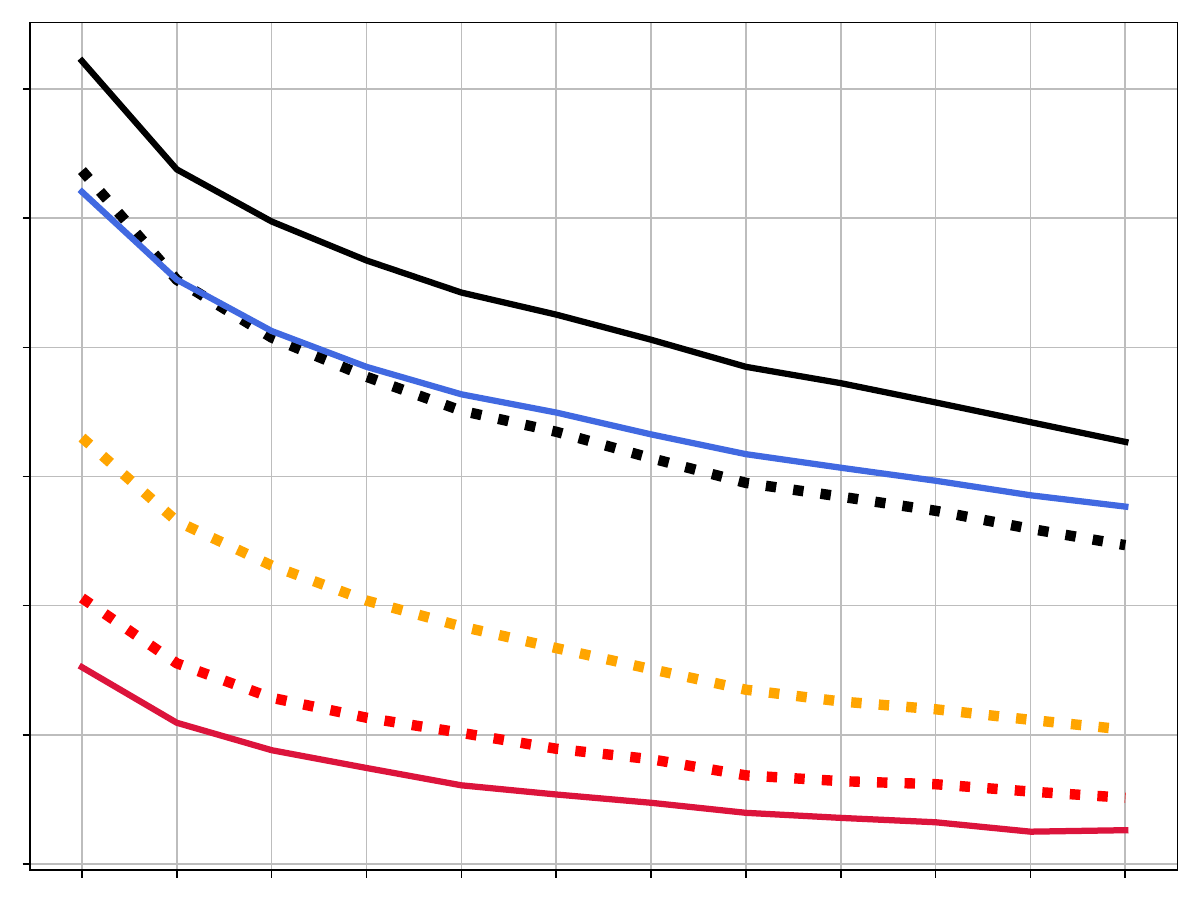}};
    \node (image2) at (5.5,-5.2) {\includegraphics[width=0.31\textwidth]{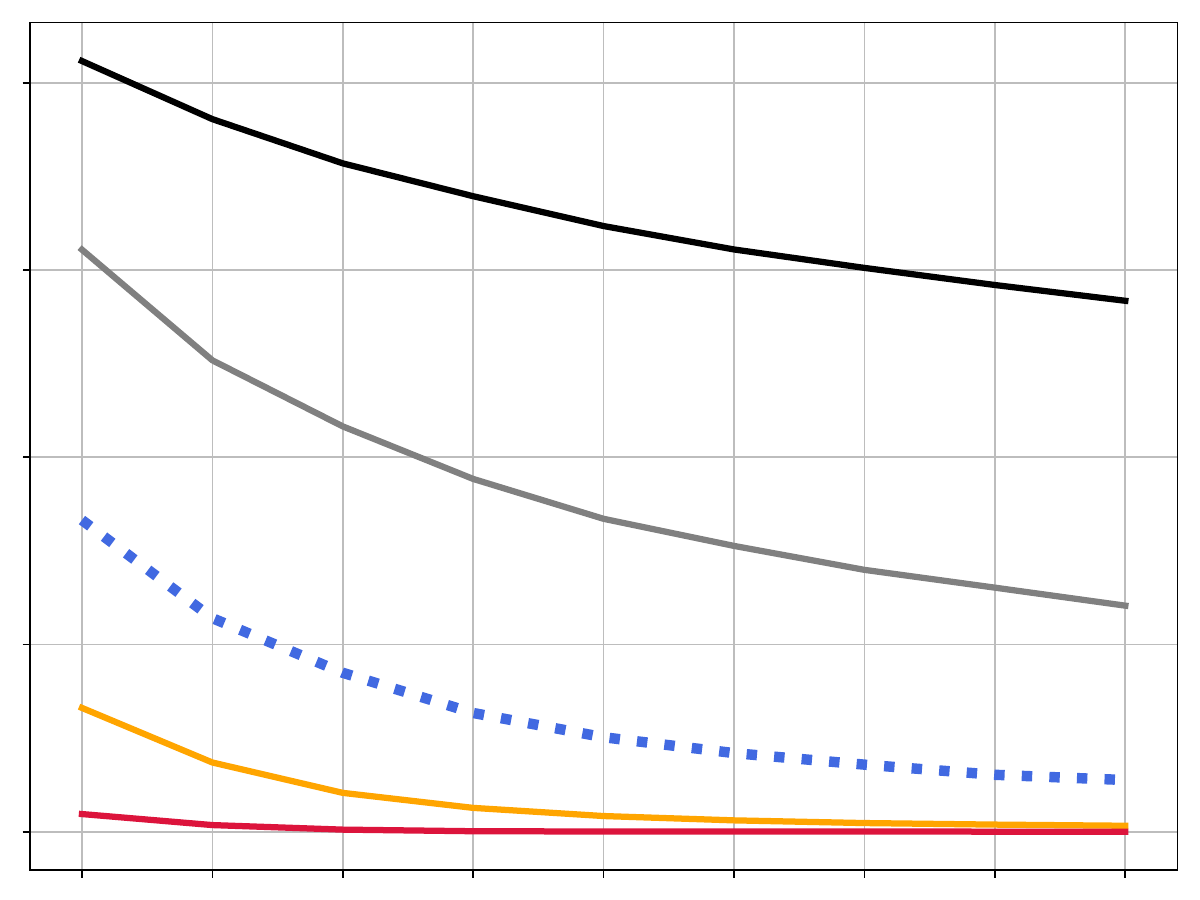}};
    \node (image2) at (11.6,-5.2) {\includegraphics[width=0.31\textwidth]{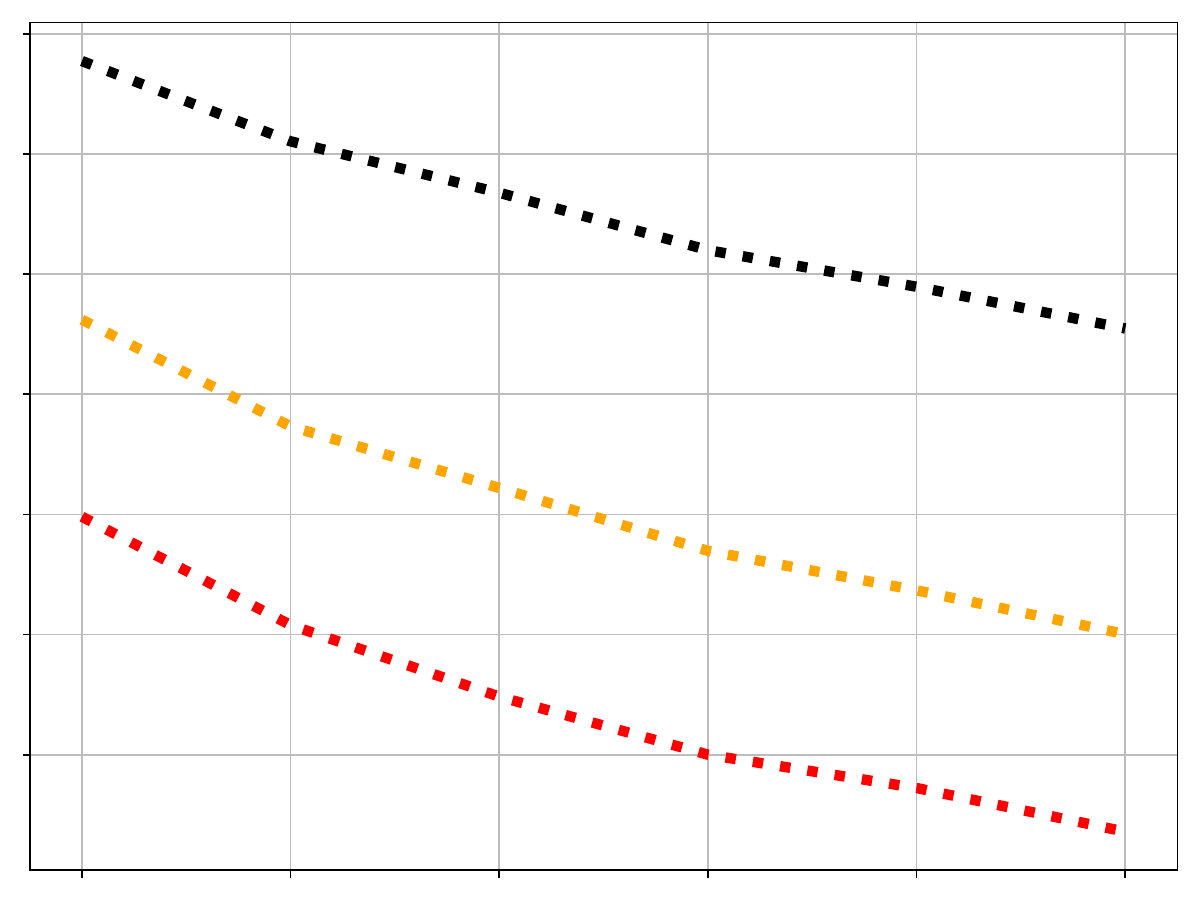}};
    \node at (0.6,-0.5) {\scriptsize{30 dBZ}}; \node at (-0.8,-1.1) {\scriptsize{\textcolor{gray}{40 dBZ}}}; \node at (-2.6,-0.6) {\scriptsize{\textcolor{blue}{50 dBZ}}};
    \node at (5.2,1.6) {\scriptsize{0.125 mm}}; \node at (5.1,0.6) {\scriptsize{1 mm}}; \node at (5.1,-0.4) {\scriptsize{\textcolor{blue}{5 mm}}}; \node at (5.2,-1.3) {\scriptsize{\textcolor{orange}{10 mm}}}; \node at (4.1,-1.6) {\scriptsize{\textcolor{purple}{15 mm}}};
    \node at (11.5,0.8) {\scriptsize{1 mm}}; \node at (11.5,-0.5) {\scriptsize{\textcolor{orange}{4 mm}}}; \node at (11,-1.1) {\scriptsize{\textcolor{red}{8 mm}}};
    \node at (-1.1,-4.2) {\scriptsize{0.2  mm}}; \node at (0.6,-5.6) {\scriptsize{1 mm}}; \node at (-2.4,-4.8) {\scriptsize{\textcolor{blue}{2 mm}}}; \node at (1,-6.25) {\scriptsize{\textcolor{orange}{4 mm}}}; \node at (-2,-6.15) {\scriptsize{\textcolor{red}{8 mm}}}; \node at (-2.6,-6.7) {\scriptsize{\textcolor{red}{20 mm}}};
    \node at (5,-3.8) {\scriptsize{0.5 mm}}; \node at (5,-5.1) {\scriptsize{\textcolor{gray}{2 mm}}}; \node at (5,-6.1) {\scriptsize{\textcolor{blue}{5 mm}}}; \node at (4.5,-6.6) {\scriptsize{\textcolor{orange}{10 mm}}}; \node at (3.45,-6.8) {\scriptsize{\textcolor{red}{30 mm}}};
    \node at (11,-3.7) {\scriptsize{1 mm}}; \node at (12,-5.4) {\scriptsize{\textcolor{orange}{4 mm}}}; \node at (11.2,-6.1) {\scriptsize{\textcolor{red}{8 mm}}};
    \node at (-0.3,-2.5) {Lead time (h)}; \node at (-0.8,-2.2) {0.5}; \node at (1.9,-2.2) {1};
    \node at (5.5,-2.5) {Lead time (h)}; \node at (5.3,-2.2) {0.5}; \node at (8,-2.2) {1};
    \node at (11.6,-2.5) {Lead time (h)}; \node at (10.6,-2.2) {0.5}; \node at (12.4,-2.2) {1}; \node at (14.1,-2.2) {1.5};
    \node at (-0.3,-7.7) {Lead time (h)}; \node at (-2.15,-7.4) {3}; \node at (-0.85,-7.4) {6}; \node at (0.55,-7.4) {9}; \node at (1.9,-7.4) {12};
    \node at (5.5,-7.7) {Lead time (h)}; \node at (6.1,-7.4) {0.5};
    \node at (11.6,-7.7) {Lead time (h)}; \node at (14.1,-7.4) {0.5};
    \node[rotate=90] at (-3.7,0) {CSI}; \node at (-3.5,-1.2) {.3}; \node at (-3.5,-0.3) {.4}; \node at (-3.5,0.5) {.5}; \node at (-3.5,1.4) {.6};
    \node[rotate=90] at (2.4,0) {CSI}; \node at (2.6,-1.8) {.0}; \node at (2.6,-1) {.2}; \node at (2.6,-0.15) {.4}; \node at (2.6,0.6) {.6}; \node at (2.6,1.45) {.8};
    \node[rotate=90] at (8.5,0) {CSI}; \node at (8.7,-1.5) {.1}; \node at (8.7,-0.8) {.2}; \node at (8.7,-0.2) {.3}; \node at (8.7,0.35) {.4}; \node at (8.7,0.95) {.5}; \node at (8.7,1.55) {.6};
    \node[rotate=90] at (-3.7,-5.2) {CSI}; \node at (-3.5,-7.2) {.0}; \node at (-3.5,-6.6) {.1}; \node at (-3.5,-5.95) {.2}; \node at (-3.5,-5.3) {.3}; \node at (-3.5,-4.75) {.4};  \node at (-3.5,-4.1) {.5}; \node at (-3.5,-3.5) {.6};
    \node[rotate=90] at (2.4,-5.2) {CSI}; \node at (2.6,-7.) {.0}; \node at (2.6,-6.1) {.2}; \node at (2.6,-5.2) {.4}; \node at (2.6,-4.35) {.6}; \node at (2.6,-3.45) {.8};
    \node[rotate=90] at (8.5,-5.2) {CSI}; \node at (8.7,-6.6) {.2}; \node at (8.7,-6.05) {.3}; \node at (8.7,-5.5) {.4}; \node at (8.7,-4.9) {.5}; \node at (8.7,-4.35) {.6}; \node at (8.7,-3.8) {.7};  \node at (8.7,-3.25) {.8};
\end{tikzpicture}  
\caption{Comparison of CSI scores ($\uparrow$) according to each threshold. The x-axis represents the time with a unit of hour, and the y-axis represents the CSI scores. \badge{N}: Non-adversarial, \badge[blue1]{A}: Adversarial, \badge[green]{U}: UNet, \badge[purple1]{D}: Diffusion, \badge[orange1]{T}: Transformer. DL models in precipitation nowcasting commonly degrade performance over time, particularly for higher thresholds. The atmosphere is a chaotic system where small perturbations in natural conditions can lead to vastly non-linear outcomes over time, a phenomenon known as the `butterfly effect'. This inherent unpredictability limits the accuracy of long-term forecasting.}
\label{tab:result2}
\end{figure*}
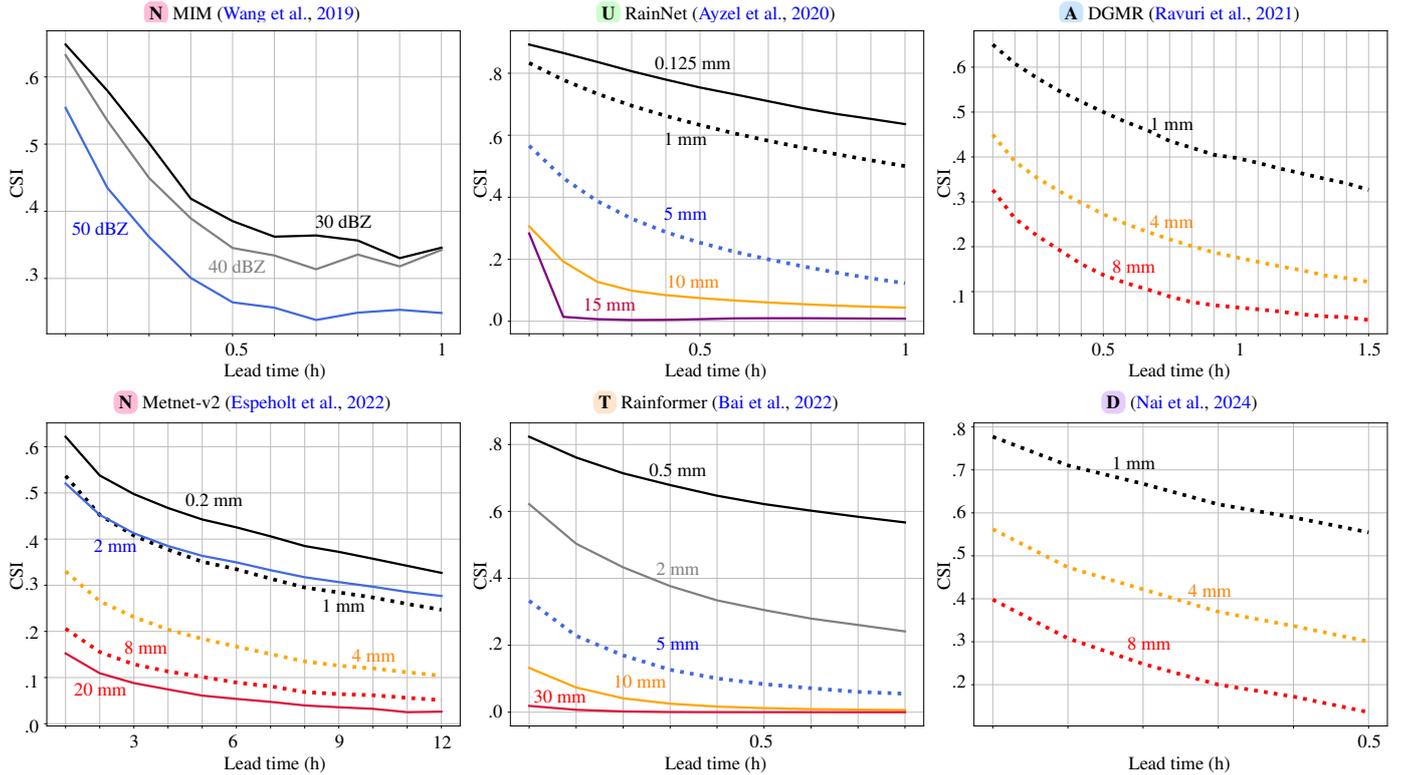

In Figure \ref{tab:result2}, we present a summary of evaluations conducted over time and across different thresholds.
Note that the datasets used are different from each other.
Based on the results, two might pose the following questions: i) How can we improve the accuracy of heavy rainfall prediction? Increasing the threshold leads to a discernible performance discrepancy, particularly in degradation for thresholds exceeding 4 mm and 40 dBZ. This outcome commonly arises due to biases in precipitation data, presenting a challenge for researchers to mitigate the data imbalance problem. ii) What approaches are available for enhancing lead time in precipitation forecasting? DL-based forecasting models demonstrate performance degradation over time, as observed in previous literature. While many models focus on forecasting rainfall within a 1 h window, the lead time presents limitations in decision-making, both socially and economically. The findings highlight the necessity for frameworks and benchmark datasets capable of developing models with longer lead times.

\section{Future challenges and opportunities}\label{sec:future}
Time series forecasting has led to discernible advancements in precipitation prediction. 
However, the full potential of DL for enhancing precipitation forecasting has yet to be investigated.
Four challenges still need to be addressed: \textcircled{1} long lead time (Section \ref{subsec:long_lead_time}), \textcircled{2} multi-sensor data fusion (Section \ref{subsec:data_fusion}), \textcircled{3} data augmentation (Section \ref{subsec:data_augmentation}) and \textcircled{4} standard evaluation protocols (Section \ref{subsec:evaluation}).
In this section, we discuss the remaining challenges and possible directions.

\subsection{Long lead time}
\label{subsec:long_lead_time}
Most studies of time series rainfall prediction using radar have focused on forecasting within a 2 h window.
The main reason for this is that predicting precipitation for more than 2 h requires high-resolution data over a wide area and significant computing resources. An observational context information radius of 72 km/h × 2 h from the target area is required to predict 2 h precipitation \citep{espeholt2022}. To incorporate a larger observational context, lowering the spatial resolution may be tempting, but it comes at the cost of losing a detailed analysis of local precipitation for the model which is not advisable. Alternative model structures may be required to address the lead time bottleneck. Full global model emulators such as FourCastNet \citep{pathak2022}, Pangu-Weather \citep{bi2022}, GraphCast \citep{lam2022}, and Gencast \citep{price2023} are capable of predicting three-dimensional atmospheric fields for medium range time scale (typically 5 -- 10 days) although precipitation forecast has not been their strong point. Their long lead time prediction is possible because their models do not have to deal with horizontal lateral boundary condition updates.

\subsection{Multi-sensor data fusion}
\label{subsec:data_fusion}
Most studies have employed radar-only models for short-term precipitation forecasting.
This choice stems from the fact that satellite data exhibit a nonlinear relationship with precipitation, making them prone to introducing noise into precipitation forecasting. 
Satellites equipped with multi-sensors offer valuable information on clouds at various altitudes, which is a crucial advantage for capturing rapidly changing precipitation phenomena. 
Thus, how can two sensor datasets complement each other? Addressing this question necessitates resolving key issues pertaining to the optimal fusion of diverse sensor datasets, including determining where and how to perform such a fusion. 
To address this effectively, ideal data fusion methods should demonstrate potent feature representations, enabling robust learning of diverse precipitation patterns by the model.

\subsection{Data augmentation}
\label{subsec:data_augmentation}
Precipitation data have imbalanced data distributions, which significantly degrade the performance of existing methods.
To address this, data augmentation procedures are required for precipitation forecasting. 
Data augmentation enhances knowledge learning by providing additional information or details, potentially improving the accuracy or reliability of the model.
\cite{seo2022b} experimented with random color jittering, random cropping, and random rotation augmentation to verify the effectiveness of data augmentation. 
They found that most procedures improved performance, but the rotation augmentation method showed a gap in the results. 
Augmenting the data in the opposite direction to the optical flow led to performance degradation.
Designing data augmentation by considering possible constraints on data characteristics can help in learning representations, thereby enhancing the performance of real-world applications in atmospheric science.

\subsection{Standard evaluation protocols}\label{subsec:evaluation}
With the increasing number of time series precipitation forecasting models, there is a growing need for a standardized baseline against which different algorithms can be uniformly evaluated. Standardizing the evaluation process is essential to ensure fair and consistent comparisons between different algorithms. In this regard, what should the test period, rainfall thresholds, and forecast lead time be? How can we compare new models to the cutting-edge models, such as different resolution models? Is there agreement on whether radar data is the best ground truth for precipitation? Perhaps an estimated value by radar is not the true rainfall, despite the radar-obtained precipitation value potentially being the closest to the actual value. By comparing the performance of different models using standardized evaluation metrics, researchers can identify the strengths and weaknesses of each approach more effectively.

\section{Conclusion}\label{sec:conclusion}
In recent years, many successful attempts have been made to develop time series models to boost precipitation forecasting. 
To compensate for the lack of a systematic summary and discourse on the practical review of precipitation forecasting, we provide a comprehensive survey of forecasting approaches while discussing the interactions and differences among them.
This survey is expected to serve as a starting point for new researchers in this area and an inspiration for future research.

\section{Acknowledgement}
This work was carried out through the R\&D project ``Development of a Next-Generation Operational System by the Korea Institute of Atmospheric Prediction Systems (KIAPS)", funded by the Korea Meteorological Administration (KMA2020-02213).
This work was supported by Institute of Information \& communications Technology Planning \& Evaluation (IITP) grant funded by the Korea government(MSIT) (No. RS-2019-II190079,  Artificial Intelligence Graduate School Program(Korea University)).

\bibliographystyle{elsarticle-harv} 
\bibliography{bibliography}

\end{document}